\documentclass[sigconf,nonacm,10pt]{acmart}
\usepackage{graphicx}   
\usepackage{subcaption}
\usepackage{pifont}
\usepackage{subcaption}
\newcommand{\cmark}{\ding{51}} % check mark
\newcommand{\xmark}{\ding{55}} % cross mark

\usepackage[utf8]{inputenc}
\usepackage[bottom]{footmisc}
\usepackage{amsmath}
\usepackage{caption}
\usepackage{subcaption}
\usepackage{mathtools} 
\usepackage{tablefootnote}
\usepackage[euler]{textgreek}
\usepackage{xcolor}
\usepackage{xcolor}
\usepackage{tikz}
\usepackage{graphicx}  % 在导言区添加
\usepackage{booktabs}
\usepackage{multirow}
\usepackage[table]{xcolor}
\usepackage{colortbl}
\usepackage[most]{tcolorbox}
\usepackage{enumitem}
\definecolor{systemframe}{HTML}{558B2F}  
\definecolor{systemback}{HTML}{C5E1A5}  
\definecolor{userframe}{HTML}{558B2F}  
\definecolor{userback}{HTML}{C5E1A5}  
\usepackage{adjustbox}

\AtBeginDocument{%
  \providecommand\BibTeX{{%
    \normalfont B\kern-0.5em{\scshape i\kern-0.25em b}\kern-0.8em\TeX}}}

\newcommand{\systemname}{KLDrive}
\usepackage{xcolor}

\author{
  Ye Tian\textsuperscript{1},
  Jingyi Zhang\textsuperscript{1},
  Zihao Wang\textsuperscript{1},
  Xiaoyuan Ren\textsuperscript{1},\\
  Xiaofan Yu\textsuperscript{2},
  Onat Gungor\textsuperscript{1},
  Tajana Rosing\textsuperscript{1}
}

\affiliation{%
  \institution{%
    \textsuperscript{1}University of California San Diego, La Jolla, California, USA\\
    \textsuperscript{2}University of California Merced, Merced, California, USA}
  \country{}
}

\email{{yet002, jiz290, ziw140, x6ren, ogungor, tajana}@ucsd.edu,  xiaofanyu@ucmerced.edu}

\begin{document}

\title{KLDrive: Fine-Grained 3D Scene Reasoning for Autonomous Driving based on Knowledge Graph}

%\author{Ye Tian, Onat Gungor, Xiaofan Yu, Tajana Rosing}
%\affiliation{%
  %\department{Computer Science and Engineering}
  %\institution{University of California San Diego}
  %\city{La Jolla}
  %\state{California}
  %\country{USA}}
%\email{{yet002, ogungor, x1yu, tajana}@ucsd.edu}

%\renewcommand{\shortauthors}{Tian, et al.}

%that make them unsuitable for deployment on onboard platforms
\begin{abstract}
Autonomous driving requires reliable reasoning over fine-grained 3D scene facts. Fine-grained question answering over multi-modal driving observations provides a natural way to evaluate this capability, yet existing perception pipelines and driving-oriented large language model (LLM) methods still suffer from unreliable scene facts, hallucinations, opaque reasoning, and heavy reliance on task-specific training. We present {\systemname}, the first knowledge-graph-augmented LLM reasoning framework for fine-grained question answering in autonomous driving. {\systemname} addresses this problem through designing two tightly coupled components: an energy-based scene fact construction module that consolidates multi-source evidence into a reliable scene knowledge graph, and an LLM agent that performs fact-grounded reasoning over a constrained action space under explicit structural constraints. By combining structured prompting with few-shot in-context exemplars, the framework adapts to diverse reasoning tasks without heavy task-specific fine-tuning. 
Experiments on two large-scale autonomous-driving QA benchmarks show that {\systemname} outperforms prior state-of-the-art methods, achieving the best overall accuracy of 65.04\% on NuScenes-QA and the best SPICE score of 42.45 on GVQA.
On counting, the most challenging factual reasoning task, it improves over the strongest baseline by 46.01 percentage points, demonstrating substantially reduced hallucinations and the benefit of coupling reliable scene fact construction with explicit reasoning.
\end{abstract}
%

% of this approach
%Additionally, by designing effective feature extraction and prompt engineering strategies, it fine-tunes the LLM, significantly improving accuracy and inference efficiency. 

%Detailed activity logs are crucial for health monitoring and personalized interventions, especially for elderly individuals with cognitive decline. Traditional methods rely on manual editing or raise privacy concerns due to the use of camera recordings. A recent study explores the possibility of using large language models (LLMs) to automatically generate life logs from sensor data. However, this approach fails to incorporate critical contextual information and physiological indicators, and it relies on a computationally intensive model. This work proposes \systemname, a system for automatically generating fine-grained, contextualized activity logs using a diverse set of sensors on smartphones and smartwatches. By designing effective feature extraction and prompt engineering strategies, we fine-tune the LLM to enhance inference efficiency. Preliminary results demonstrate the feasibility of the proposed approach. 

%\ccsdesc[500]{Computing methodologies~Machine learning}
%\ccsdesc[300]{Computer systems organization~cyber-physical systems}


%``''
\maketitle

\section{Introduction}

Autonomous driving relies on accurate understanding of fine-grained scene facts, including object identities, motion states, and spatial relations~\cite{shi2025motion,zhao2025survey}. These facts directly govern high-level decisions~\cite{xia2024survey}, such as braking, acceleration, lane changes, and turning maneuvers. Reliable reasoning over the 3D driving scene therefore forms a critical interface between perception and planning.

Fine-grained question answering (QA) over driving scenes provides a natural and effective testbed for evaluating whether a system can understand fine-grained scene facts~\cite{qian2024nuscenes,marcu2024lingoqa,sima2024drivelm,xu2024drivegpt4}. Given a temporal window of multi-modal sensor observations, primarily multi-view cameras and LiDAR, together with a factual query about the current scene, the system is expected to understand and answer questions about object existence, motion states, counts, and spatial relations. For example, it may need to count how many vehicles are approaching the ego vehicle within a specified distance range, or determine whether there exists a moving target in a particular relative direction of the ego vehicle. Unlike conventional perception tasks that primarily evaluate recognition or localization accuracy, fine-grained QA instead tests whether the system can recover and reason over the structured scene facts that truly matter for driving. Therefore, it exposes weaknesses along the perception-to-reasoning pipeline, thereby offering more targeted evidence for model diagnosis, system improvement, and reliability assessment in safety-critical scenarios.

\begin{figure}[t]
    \centering
    \includegraphics[width=0.44\textwidth]{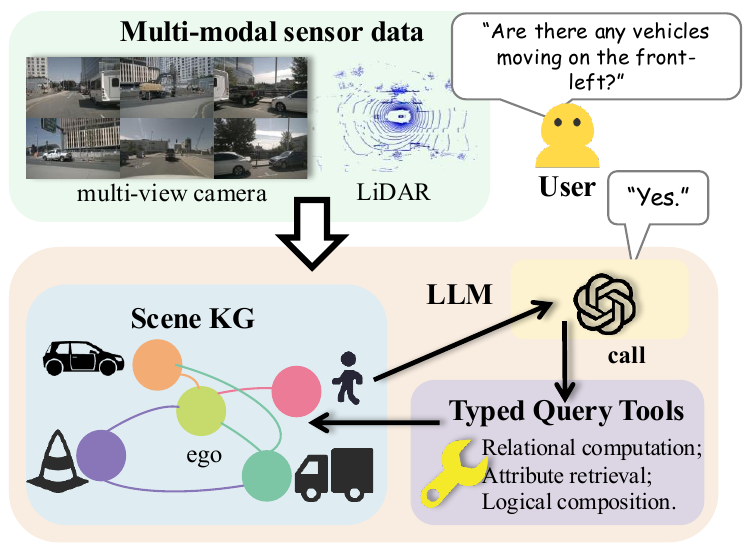} % Adjust width to column width
    %\vspace{-1mm}
    \vspace{-0.4cm}
    \caption{\systemname: fact-based scene question answering via a 3D scene KG and a tool-constrained LLM agent.}
    \label{fig:firstpicture}
    \vspace{-0.6cm}
\end{figure}

In this context, we focus on an important problem that remains insufficiently addressed: given multi-modal temporal sensor data, how can a system understand a complex driving scene and perform reliable reasoning? Notably, we study such fine-grained complex reasoning in an offline analysis setting. At the current stage, we primarily focus on the accuracy of the reasoning results and the reliability of the reasoning process, rather than millisecond-level online response speed. Once these more fundamental issues are better addressed, efficiency optimization for real-time deployment can be pursued as an important direction for future work.

%%%%%%%%%%%%%%%%%%%%%
%challenge:
Achieving this goal presents three key challenges.
\textbf{(1) Reliable scene fact construction.}
Fine-grained question answering relies on accurate object identities, motion states, and spatial relations. However, recovering such scene facts from multi-view camera and LiDAR streams is difficult due to illumination variations, occlusion, ego motion, and sensor noise. Even state-of-the-art detectors still produce missed detections, false positives, duplicate candidates, and inconsistent hypotheses, making their outputs unreliable as a direct basis for reasoning.
\textbf{(2) Fact-grounded interpretable reasoning.}
Autonomous driving is a safety-critical domain, which requires system answers to be supported by concrete and verifiable scene facts. However, existing end-to-end methods directly map inputs to answers in a black-box manner, resulting in opaque decision processes and hallucinations. Their outputs also cannot be traced back to specific and verifiable scene evidence.
\textbf{(3) Task adaptation under limited supervision.}
Fine-grained scene question answering spans diverse query forms and reasoning patterns, yet collecting large-scale annotated question--program pairs and conducting task-specific training for each deployment setting is costly and difficult to scale. Therefore, the system must efficiently adapt to diverse reasoning tasks under limited supervision, without relying on heavy task-specific fine-tuning pipelines.
%%%%%%%%%%%%%%%%%%%%%%%%%

Existing approaches still fall short of addressing the above challenges. As shown in Table~\ref{tab:sota_comparison}, no representative method simultaneously provides strong perception ability, reasoning ability, freedom from task-specific training, and explainable reasoning.
First, perception-based methods, such as RayDN~\cite{liu2024ray}, FocalFormer3D~\cite{chen2023focalformer3d}, and IS-Fusion~\cite{yin2024fusion}, possess strong scene perception ability, but still lack the explicit reasoning capability required for fine-grained factual question answering.
Second, large language models (LLMs) have demonstrated strong capability on general reasoning tasks~\cite{peng2024multimath,wang2024measuring,xu2025vs}, but remain prone to hallucinations in fine-grained spatial and factual reasoning~\cite{wang2024mitigating,liu2024survey,li2025fine}. To improve adaptation to autonomous driving scenarios, recent works have proposed driving-oriented LLM frameworks, such as DriveLM~\cite{sima2024drivelm}, LiDAR-LLM~\cite{yang2025lidar}, MAPLM~\cite{cao2024maplm}, and CREMA~\cite{yu2025crema}, by introducing task-specific encoders and training on large-scale driving data, representing the current state-of-the-art (SOTA) among LLM-based methods. However, as shown by our experiments, these methods still suffer from severe hallucinations on fine-grained factual queries over 3D scenes. In addition, their reliance on task-specific training pipelines makes it difficult to efficiently adapt to diverse reasoning tasks under limited supervision.
Moreover, these methods lack fact-grounded and interpretable reasoning, instead relying on end-to-end black-box mappings from multi-modal inputs to answers. Consequently, their predictions cannot be traced to concrete and verifiable facts, which is highly risky for safety-critical autonomous driving.

%%%%%%%%%%%%%%%%%
\begin{table}[t]
  \centering
  \renewcommand{\arraystretch}{1.1}
  \scriptsize
  \resizebox{0.46\textwidth}{!}{%
  \begin{tabular}{l l c c c c}
    \toprule
    & Methods &
    \shortstack{Perception\\ability} &
    \shortstack{Reasoning\\ability} &
    \shortstack{No task-specific\\training} &
    \shortstack{Explainable\\reasoning} \\
    \midrule
    % ------- 3D perception -------
    \multirow{3}{*}{ \shortstack{Perception-based \\Methods}}
      & \cellcolor{gray!15}RayDN~\cite{liu2024ray}
      & \cellcolor{gray!15}\cmark
      & \cellcolor{gray!15}\xmark
      & \cellcolor{gray!15}\xmark
      & \cellcolor{gray!15}\xmark \\

      & FocalFormer3D~\cite{chen2023focalformer3d}
      & \cmark & \xmark & \xmark & \xmark \\

      & \cellcolor{gray!15}IS-Fusion~\cite{yin2024fusion}
      & \cellcolor{gray!15}\cmark
      & \cellcolor{gray!15}\xmark
      & \cellcolor{gray!15}\xmark
      & \cellcolor{gray!15}\xmark \\
    \midrule
    % ------- Driving-oriented -------
    \multirow{4}{*}{\shortstack{Driving-oriented\\LLM frameworks}}
      & \cellcolor{gray!15}DriveLM~\cite{sima2024drivelm}
      & \cellcolor{gray!15}\cmark
      & \cellcolor{gray!15}\cmark
      & \cellcolor{gray!15}\xmark
      & \cellcolor{gray!15}\xmark \\

      & LiDAR-LLM~\cite{yang2025lidar}
      & \cmark & \cmark & \xmark & \xmark \\

      & \cellcolor{gray!15}MAPLM~\cite{cao2024maplm}
      & \cellcolor{gray!15}\cmark
      & \cellcolor{gray!15}\cmark
      & \cellcolor{gray!15}\xmark
      & \cellcolor{gray!15}\xmark \\

      & CREMA~\cite{yu2025crema}
      & \cmark & \cmark & \xmark & \xmark \\
    \midrule
     \qquad Ours & \cellcolor{gray!15}{\systemname} 
      & \cellcolor{gray!15}\cmark & \cellcolor{gray!15}\cmark & \cellcolor{gray!15}\cmark & \cellcolor{gray!15}\cmark \\
    \bottomrule
  \end{tabular}}
\caption{Comparison of existing representative methods and \systemname.
  \cmark\ indicates the property is supported and \xmark\ indicates it is not.}
\label{tab:sota_comparison}
  \vspace{-0.8cm}
\end{table}

%%%%%%%%%%%%%%%%%

\begin{figure*}[t]
    \centering
    \includegraphics[width=0.9\textwidth]{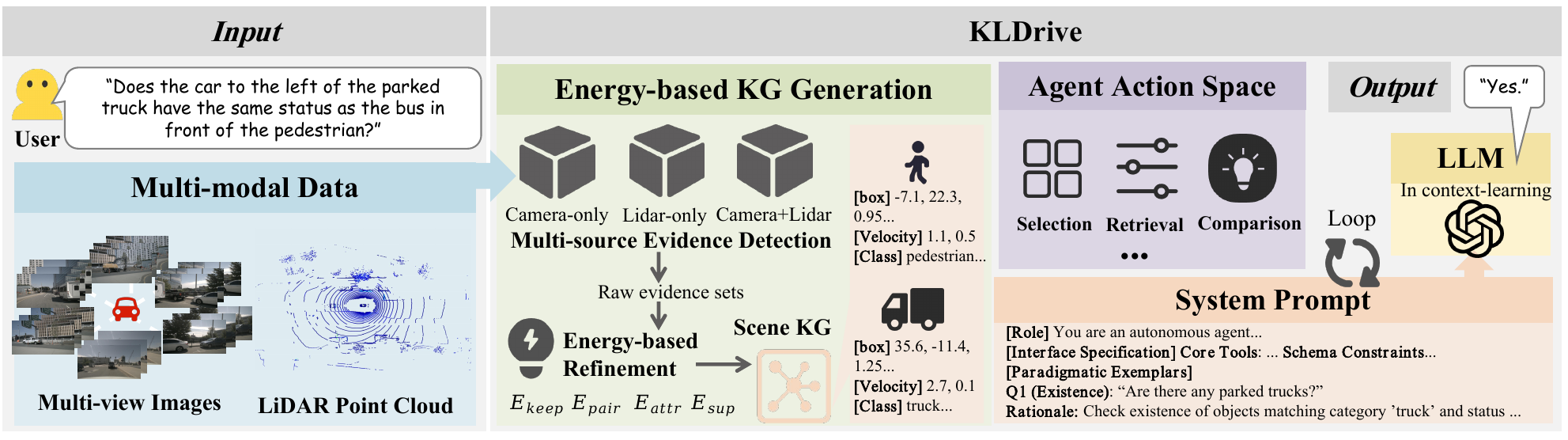} % Adjust width to column width
    \vspace{-0.3cm}
    \caption{Overview of \systemname, which converts multi-modal sensor data and a natural-language query into a scene KG, applies generic reasoning tools, and uses an in-context–learned LLM agent to produce a fact-grounded answer.}
    \label{fig:systemoverview}
    \vspace{-0.2cm}
\end{figure*}

To bridge these gaps, we propose \systemname, the first knowledge graph (KG)-augmented LLM reasoning framework for fine-grained question answering in autonomous driving. As shown in Figure~\ref{fig:systemoverview}, {\systemname} designs two tightly coupled modules: reliable scene fact construction and an LLM agent for reasoning over scene facts. Specifically, \textbf{(1)} to recover reliable scene facts from noisy driving observations, {\systemname} first formulates perception as a multi-source evidence construction problem. It consolidates complementary hypotheses from camera and LiDAR point clouds through cross-source pooling and temporal recovery, and proposes an energy model to resolve conflicts, refine a coherent and reliable set of scene entities, and organize them into the scene KG.
\textbf{(2)} Built upon this scene KG, we design an LLM agent that plans over a constrained action space to perform fact-grounded and interpretable reasoning. Rather than directly generating answers, the LLM acts as a semantic planner that incrementally invokes tools and receives feedback under a Plan--Execute--Observe loop. Each action is executed on concrete scene entities, thereby turning the reasoning into an explicit and auditable process grounded in scene facts.
\textbf{(3)} In addition, to support task adaptation under limited supervision, {\systemname} avoids task-specific program supervision and heavy fine-tuning pipelines. Through the design of a constrained action space together with few-shot in-context exemplars, it guides the LLM to invoke different tools under explicit rules and constraints, thereby adapting to diverse reasoning tasks.

We implement {\systemname} with Qwen3-7B as the LLM backbone without task-specific fine-tuning and conduct systematic evaluations on two large-scale fine-grained autonomous driving QA benchmarks, NuScenes-QA~\cite{qian2024nuscenes} and GVQA~\cite{sima2024drivelm}. Extensive experiments show that {\systemname} achieves the best performance on both benchmarks. On the structured QA task of NuScenes-QA~\cite{qian2024nuscenes}, {\systemname} attains the best overall accuracy of 65.04\%, surpassing the strongest baseline at 60.17\%; on GVQA~\cite{sima2024drivelm}, it achieves the best SPICE score of 42.45. When the perceptual facts are fully correct, its overall accuracy further reaches 84.49\%. More importantly, on counting, the most challenging fine-grained factual reasoning task, {\systemname} achieves an accuracy of 64.46\%, improving over the strongest baseline by 46.01 points, demonstrating a significant advantage in mitigating hallucinations of end-to-end methods.

%%%%%%%%%%%%%%%%

\section{How to Construct Reliable Scene KG?}
%How to Construct Reliable Scene KG?
%for Reliable Scene KG Construction

\subsection{Challenges}
Fine-grained reasoning in 3D autonomous driving scenes requires more than recognizing isolated objects. It relies on a reliable scene representation that faithfully captures object identities, attributes, motion states, and spatial relations, as these structured scene facts form the basis of reasoning. Consequently, accurate perception and high-quality scene knowledge graph construction are prerequisites for reliable fine-grained scene understanding. However, building such a knowledge graph from real-world driving sensor streams remains highly challenging.

\textbf{Challenge 1: Imperfect Evidence Acquisition.}
Real-world driving scenes are affected by occlusion, illumination variation, sensor noise, and modality-specific degradation, making it difficult for any single sensing pathway to produce uniformly reliable scene facts. Cameras provide rich semantic textures and visual cues that are critical for object recognition, but they are sensitive to lighting conditions. LiDAR provides accurate 3D geometry and depth measurements and is relatively robust to illumination changes, but it suffers from long-range sparsity and degraded reliability for distant objects. Fusion-based detectors often provide a better balance between semantic and geometric cues, yet their outputs are corrupted by cross-modal misalignment, synchronization noise, or fusion bias. As a result, directly constructing a scene graph from the hard decisions of a single detector is fragile. A more robust strategy is to treat perception as a multi-source evidence acquisition problem and aggregate complementary hypotheses from heterogeneous detectors into a unified candidate space.

\textbf{Challenge 2: Structured Ambiguity.}
Even after aggregating multi-source evidence, the resulting candidate space still contains structured fuzziness rather than just independent noise. Some candidates correspond to redundant hypotheses of the same physical object across detectors, some are weakly observed in the current frame but supported by neighboring frames, and some may appear locally plausible in isolation while remaining incompatible with the global scene interpretation. Therefore, recovering the final scene entities cannot be reduced to simple score thresholding or independent per-candidate classification. Instead, it requires structured inference over the entire candidate set, where the selected entities should jointly satisfy strong evidence support, low redundancy, temporal consistency, and semantic plausibility.

\subsection{EBM Background}

Energy-based models (EBMs)~\cite{lecun2006tutorial,xu2024energy} provide a framework for structured prediction by associating a scalar energy with each input--output configuration, where lower energy indicates higher compatibility between the observed evidence and the predicted output. Given observed evidence $\mathbf{x}$ and a structured output space $\mathcal{Y}(\mathbf{x})$, inference takes the form:
\begin{equation}
\mathbf{y}^{*}
=
\arg\min_{\mathbf{y}\in\mathcal{Y}(\mathbf{x})}
E_{\theta}(\mathbf{y}, \mathbf{x}),
\label{eq:ebm_generic}
\end{equation}
where $E_{\theta}$ is a learnable energy function and $\mathbf{y}^{*}$ denotes the lowest-energy output under the observed evidence. In this way, EBM does not make isolated local decisions for individual hypotheses, but evaluates the compatibility of a structured output as a whole.

This mechanism is particularly suitable for our problem. In scene entity recovery, the goal is not to decide whether each candidate should be retained independently, but to infer a subset of scene entities that is jointly well supported by the available evidence and coherent at the scene level. The final decision should account for multiple heterogeneous factors, including detector confidence, cross-source corroboration, pairwise redundancy, temporal support, and attribute-level consistency. EBMs provide a principled way to integrate these factors into a unified refinement objective, so that the final output is determined by the overall quality of the recovered scene interpretation.

However, applying EBM refinement to driving scenes is nontrivial. The energy function must reconcile heterogeneous detector outputs, distinguish harmful redundancy from useful temporal corroboration, and encode semantic and physical consistency while remaining tractable for inference over a large combinatorial output space. These requirements make the refinement stage both essential and challenging. To address these challenges, we design an energy-based evidence refinement framework that jointly evaluates support, interaction, temporal consistency, and semantic plausibility over pooled multi-source and multi-frame candidates, and recovers a refined set of reliable scene entities for subsequent scene KG construction.

\section{{\systemname} Overview}
{\systemname} is built on a crucial design principle: complex and fine-grained scene question answering requires both a trustworthy scene representation and a reliable reasoning process. To this end, as illustrated in Figure~\ref{fig:systemoverview}, the system is organized into two tightly coupled stages. The first stage converts noisy multi-modal driving observations into a reliable scene KG by consolidating complementary multi-source evidence, refining candidate scene entities with an energy-based model, and organizing the refined entities into structured scene facts. The second stage performs question answering over this KG using an LLM agent that plans and executes grounded symbolic operations under a bounded action space. By coupling energy-based scene KG generation with LLM-driven orchestration and reasoning, {\systemname} produces accurate, interpretable answers for complex 3D driving scenes.

\section{Energy-based Knowledge Graph Generation}
\label{sec:ebm-KG}

%As discussed above, directly constructing a reliable scene KG from raw driving sensor streams is difficult due to imperfect sensor evidence and structured ambiguity. To address this, {\systemname} generates the scene KG in three stages, as illustrated in Fig.~\ref{fig:perception-ebm}. It first consolidates complementary multi-source evidence through cross-source pooling and temporal recovery to form a unified candidate space, then proposed an energy-based model to refine reliable scene entities, and finally organizes the refined entities into a scene KG for subsequent LLM-driven reasoning.

%As illustrated in Figure~\ref{fig:systemoverview},
\subsection{Multi-source Evidence Detection.}
To build reliable scene facts, {\systemname} first acquires object evidence from synchronized multi-view images and LiDAR point clouds.
Since real-world driving scenes are affected by occlusion, illumination variation, sensor noise, and modality-specific degradation, no single sensing pathway can provide uniformly reliable evidence.
As illustrated in Figure~\ref{fig:systemoverview}, we therefore formulate perception as a multi-source evidence acquisition problem and collect complementary hypotheses from camera-based, LiDAR-based, and fusion detectors. 
In our implementation, the detector set is instantiated with RayDN~\cite{liu2024ray}, FocalFormer3D~\cite{chen2023focalformer3d}, and IS-Fusion~\cite{yin2024fusion}, selected for their strong empirical performance and complementary sensing characteristics.
They together provide semantically rich visual evidence, geometrically grounded LiDAR evidence, and cross-modal corroborative evidence.

Formally, let $\mathcal{M}$ denote the detector set. For frame $\tau$, each detector $m\in\mathcal{M}$ processes its corresponding modality subset and produces a raw evidence set
\[
\mathcal{Y}_\tau^{m}
=
\left\{
\left(\mathbf{b}_{i}^{\tau,m},\, c_{i}^{\tau,m},\, s_{i}^{\tau,m}\right)
\right\}_{i=1}^{M_{\tau,m}},
\]
where $\mathbf{b}_{i}^{\tau,m}$ denotes the object state, $c_{i}^{\tau,m}$ the semantic class label, and $s_{i}^{\tau,m}\in[0,1]$ the detector confidence. The union of all $\mathcal{Y}_\tau^{m}$ forms a heterogeneous raw evidence space, which is then consolidated into a unified candidate space by the consensus-aware pooling stage.

%%%%%%%%%%%%%%%%%%%%%%%
\begin{figure}[t]
    \centering
    \includegraphics[width=0.49\textwidth]{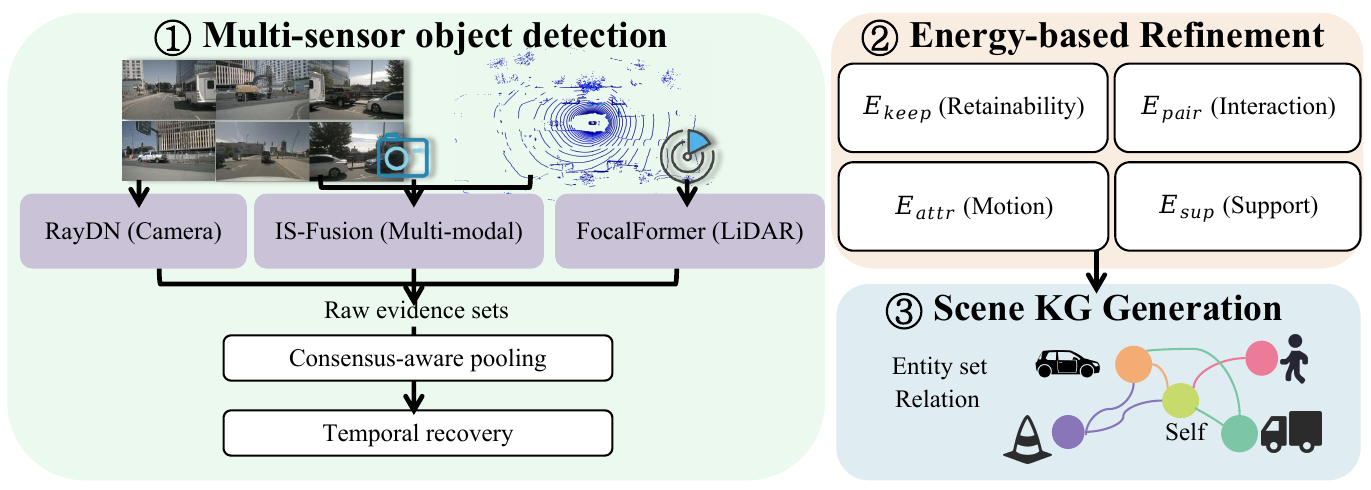} % Adjust width to column width
    %\vspace{-1mm}
    \vspace{-0.6cm}
\caption{Energy-based scene KG generation model.}
    \label{fig:perception-ebm}
    \vspace{-0.3cm}
\end{figure}
%%%%%%%%%%%%%%%%%%%%%%%

\subsection{Evidence Pooling and Recovery}

Although the detectors provide complementary evidence, their outputs remain heterogeneous in geometric bias and confidence scale, and thus cannot yet form a unified evidence space. Moreover, real objects may still be weakly supported or entirely missing in the current frame due to occlusion, motion blur, or temporary sensing degradation. 
We therefore combine consensus-aware cross-source pooling with temporal recovery over neighboring frames to form a unified evidence space with explicit temporal support for the subsequent energy-based refinement stage.

\textbf{Evidence Pooling.}
We first perform consensus-aware pooling over same-frame detector outputs. All raw evidence items are transformed into a shared ego-centric coordinate frame, and detector-specific confidence scores are normalized to reduce cross-source scale mismatch. We then associate hypotheses from different detectors when they share the same semantic label and remain geometrically consistent in BEV position, size, and heading. 
Each associated group is aggregated into one pooled candidate $\tilde{y}_i^\tau$, whose 3D state, semantic label, and confidence score are obtained by aggregating its supporting evidence items.
Importantly, the supporting detector identities are explicitly retained in a provenance set $\mathcal{M}_i^\tau \subseteq \mathcal{M}$, allowing the subsequent refinement stage to determine which detectors support each candidate and how strong that cross-source corroboration is.

\textbf{Temporal Recovery.}
While cross-source pooling improves evidence consistency within the current frame, the resulting candidate space may still miss real objects that are only weakly observed or temporarily absent due to short-term occlusion, motion blur, or transient sensing degradation. To address this limitation, we further incorporate evidence from neighboring frames around $\tau$ and recover missing object states through short-horizon temporal association and interpolation. Specifically, we associate pooled candidates across a short temporal window to form tracklets in the global frame, and perform trajectory-based completion over valid gaps between two observed states. For a tracklet with observed centers $\mathbf{p}_i^{\tau^-}$ and $\mathbf{p}_i^{\tau^+}$ at two neighboring frames, the missing center at an intermediate frame $\tau$ is recovered by linear interpolation:
\begin{equation}
\tilde{\mathbf{p}}_i^\tau
=
(1-\alpha)\mathbf{p}_i^{\tau^-}
+
\alpha \mathbf{p}_i^{\tau^+},
\qquad
\alpha
=
\frac{\tau-\tau^-}{\tau^+-\tau^-}\in[0,1].
\label{eq:temporal_interp}
\end{equation}
The recovered center is then projected back to the local sensor frame at time $\tau$, while box attributes and semantic labels are inherited from the nearest observed state along the tracklet. Each recovered hypothesis is assigned the average confidence of its two neighboring observations and merged into the candidate pool.

\textbf{Evidence Space.}
After cross-source pooling and temporal recovery, we obtain the final pooled candidate set
$\widetilde{\mathcal{Y}}_\tau$=$\{\tilde{y}_i^\tau\}_{i=1}^{K_\tau}$,
where each candidate $\tilde{y}_i^\tau$ includes its aggregated 3D state, semantic label, pooled confidence $\tilde{s}_i^\tau$, provenance set $\mathcal{M}_i^\tau$, and an evidence-type indicator $g_i$ that distinguishes directly observed candidates from temporally recovered ones. This pooled candidate set serves as the input to the subsequent energy-based refinement stage.

\subsection{Energy Model Formulation.}
Given the pooled candidate set, {\systemname} aims to recover the final scene entities at frame $\tau$. This is challenging because the desired output cannot be obtained by evaluating each candidate independently: whether a candidate should be retained depends not only on its own evidence, but also on cross-source corroboration, interactions with other candidates, and the global coherence of the resulting scene interpretation.

Inspired by energy-based modeling, we formulate this process as a structured subset refinement problem over $\widetilde{\mathcal{Y}}_\tau$ and design a refinement energy model to favor candidate subsets that recover reliable scene entities, thereby enabling higher-quality scene KG construction.
Let $\mathbf{z}_\tau=\{z_i^\tau\}_{i=1}^{K_\tau}$ denote binary selection variables, where $z_i^\tau=1$ indicates that candidate $\tilde{y}_i^\tau$ is retained in the final scene entity set, and $z_i^\tau=0$ otherwise. We define the total refinement energy as
\begin{equation}
\begin{aligned}
E(\widetilde{\mathcal{Y}}_\tau,\mathbf{z}_\tau)
&=
\sum_i z_i^\tau E_{\mathrm{keep}}(i)
+
\sum_{i<j} z_i^\tau z_j^\tau E_{\mathrm{pair}}(i,j) \\
&\quad+
\sum_i z_i^\tau E_{\mathrm{attr}}(i)
+
\sum_i z_i^\tau E_{\mathrm{sup}}(i),
\end{aligned}
\label{eq:four_term_energy}
\end{equation}
where $E_{\mathrm{keep}}$ captures the unary retainability of each candidate, $E_{\mathrm{pair}}$ models pairwise interactions among candidates, $E_{\mathrm{attr}}$ enforces motion-state plausibility, and $E_{\mathrm{sup}}$ penalizes insufficient temporal or contextual support. The refinement objective is to optimize $\mathbf{z}_\tau$ such that the selected candidate subset attains low total energy under Eq.~\eqref{eq:four_term_energy}. Lower energy corresponds to a scene entity set with stronger evidence support, less redundancy, and better semantic and temporal consistency.

\textbf{Retainability and Interaction Modeling.}
We first model candidate-level evidence strength through the keep component, which determines whether an individual candidate is sufficiently supported to be retained. Specifically, we define
\begin{equation}
E_{\mathrm{keep}}(i)=-\,\ell_i^{\mathrm{keep}},
\qquad
\ell_i^{\mathrm{keep}}
=
a_{g_i}\,\tilde{s}_i^\tau+b_{g_i}+\alpha_{\mathrm{src}}k_i,
\label{eq:keep_energy}
\end{equation}
where $\tilde{s}_i^\tau$ is the pooled confidence score of candidate $\tilde{y}_i^\tau$, and $g_i$ denotes its evidence type, indicating whether the candidate is directly observed or temporally recovered.
The parameters $a_{g_i}$ and $b_{g_i}$ calibrate the pooled confidence according to the candidate's evidence type. 
Moreover, $k_i=|\mathcal{M}_i^\tau|/|\mathcal{M}|$ denotes the normalized cross-source corroboration strength derived from the provenance set $\mathcal{M}_i^\tau\subseteq\mathcal{M}$. Therefore, $E_{\mathrm{keep}}$ favors candidates that have both stronger calibrated confidence and stronger cross-source support.

We next model pairwise interactions among candidates through the pair component:
\begin{equation}
E_{\mathrm{pair}}(i,j)
=
\lambda_{\mathrm{dup}}\,r_{ij}^{\mathrm{dup}}
-
\lambda_{\mathrm{tmp}}\,r_{ij}^{\mathrm{tmp}},
\label{eq:pair_energy}
\end{equation}
where $r_{ij}^{\mathrm{dup}}$ measures the degree to which candidates $\tilde{y}_i^\tau$ and $\tilde{y}_j^\tau$ provide redundant explanations of the same physical object, using pairwise cues such as BEV distance, overlap, and semantic label consistency. In contrast, $r_{ij}^{\mathrm{tmp}}$ measures pairwise consistency induced by temporally supported evidence, favoring candidate pairs whose geometry and semantics remain compatible with neighboring-frame corroboration. 
Accordingly, $E_{\mathrm{pair}}$ penalizes redundant candidate pairs while preserving pairwise consistency supported by neighboring-frame corroboration.

\textbf{Semantic and Support-guided Correction.}
Beyond unary evidence strength and pairwise interaction, we further regularize candidate selection through motion-state plausibility and support sufficiency. Specifically, the attr component is defined as
\begin{equation}
E_{\mathrm{attr}}(i)=-\log p(a_i \mid v_i),
\label{eq:attr_energy}
\end{equation}
where $a_i$ denotes the motion state of candidate $\tilde{y}_i^\tau$, and $v_i$ denotes its kinematic evidence derived from the estimated speed magnitude together with its local motion context. Thus, $E_{\mathrm{attr}}$ assigns lower energy to candidates whose motion states are more consistent with their kinematic evidence, enforcing an additional physical plausibility constraint beyond geometric support alone.
We further model support sufficiency through the sup component:
$E_{\mathrm{sup}}(i)
=
\beta_{\mathrm{tmp}}\,[m_{\mathrm{tmp}}-u_i]_+
+
\beta_{\mathrm{ctx}}\,[m_{\mathrm{ctx}}-d_i]_+,
\label{eq:sup_energy}$
where $u_i$ denotes the aggregated temporal support of candidate $i$ from neighboring-frame evidence, $d_i$ denotes the local evidence density around the candidate, $m_{\mathrm{tmp}}$ and $m_{\mathrm{ctx}}$ are the corresponding minimum support thresholds, and $[x]_+=\max(x,0)$. Accordingly, $E_{\mathrm{sup}}$ assigns higher energy to candidates whose temporal corroboration from neighboring frames or local evidence density remains insufficient, thereby suppressing weakly grounded hypotheses.

%\textbf{Learning and Inference.}
Together, the four terms define a structured refinement objective over the pooled candidate set. In our implementation, the learnable parameters are estimated using binary supervision on pooled candidates, and inference optimizes Eq.~\eqref{eq:four_term_energy} over $\mathbf{z}_\tau$ to produce the refined entity set $\hat{\mathcal{Y}}_\tau$ for subsequent scene KG construction.

%Together, these four terms define a structured refinement objective. In practice, {\systemname} learns a logistic keep head with binary cross-entropy on the pooled candidate set, and at inference time, its learned keep logits are incorporated into the four-term energy and optimized via a staged dual-head refinement procedure, yielding the refined scene entities for subsequent scene graph construction.
%Together, the four components enable refinement to jointly model candidate retainability, pairwise interaction, motion plausibility, and support sufficiency, thereby favoring scene entity sets that are well supported, non-redundant, and temporally coherent.

\subsection{Scene Knowledge Graph Generation}
\label{sec:KGGeneration}

Given the refined entity set $\hat{\mathcal{Y}}_\tau$ obtained after energy-based refinement, we next construct a scene knowledge graph for structured LLM reasoning. At frame $\tau$, each refined entity is instantiated as a node $v_i\in\mathcal{V}_\tau$ with attribute vector $\mathbf{a}_i^\tau$, which encodes its geometry, motion state, semantic label, and confidence. In this way, the refined candidate subset is converted into an object-centric symbolic scene representation.

Rather than explicitly materializing all pairwise edges, we represent inter-object relations through a compact library of relational operators $\Psi$ defined over node attributes. For any pair of nodes $(v_i,v_j)$ and operator $\psi\in\Psi$, the corresponding relation is computed as: $\mathbf{r}_{ij,\psi}^\tau=\psi(\mathbf{a}_i^\tau,\mathbf{a}_j^\tau)$.
For example, a basic geometric operator produces the relative displacement and distance between two entities:
\[
\mathbf{r}_{ij,\mathrm{geo}}^\tau=
[\Delta x_{ij},\Delta y_{ij},\Delta z_{ij},\|\Delta\mathbf{p}_{ij}\|_2]^\top.
\]
These operator-defined relations provide the basis for higher-level query-facing reasoning over spatial, semantic, and motion-aware scene structure.
Since full pairwise relation materialization incurs $O(|\mathcal{V}_\tau|^2)$, {\systemname} instantiates only the relations relevant to the queried objects and reasoning operators. This query-adaptive construction preserves the structured scene information needed for reasoning while avoiding unnecessary calculation, supporting the subsequent LLM-driven orchestration and reasoning stage.

\begin{figure}[h]
    \centering
    \includegraphics[width=0.48\textwidth]{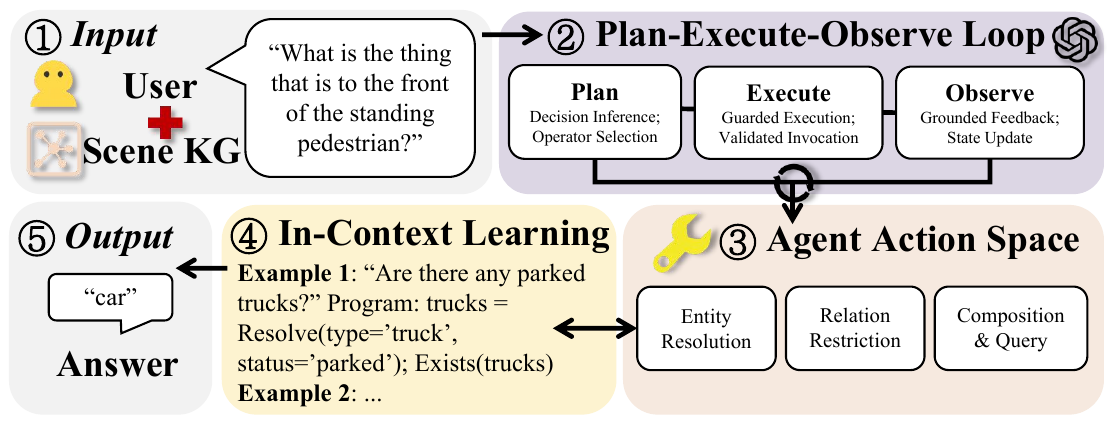} % Adjust width to column width
    %\vspace{-1mm}
    \vspace{-0.6cm}
    \caption{Proposed LLM Scene-Reasoning Agent.}
    \label{fig:freasoning_overview}
    \vspace{-0.5cm}
\end{figure}

\section{LLM-Driven Orchestration and Reasoning}
\label{sec:llm-agent}
Built on the constructed scene KG, we design an LLM agent for reliable fine-grained scene reasoning. As illustrated in Figure~\ref{fig:freasoning_overview}, the agent operates over a bounded scene-query algebra defined on the scene KG, while the LLM serves as a semantic planner that selects executable operator calls to analyze complex queries step by step rather than directly generating answers. The overall process follows a \emph{Plan--Execute--Observe} loop: the planner proposes an action, the executor applies it to the scene KG, and the grounded observation is fed back into subsequent planning. This design turns scene reasoning into an explicit and auditable reasoning process over structured scene facts.

\subsection{The Plan--Execute--Observe Loop.}
We design an LLM agent that realizes question answering step by step through a \emph{Plan--Execute--Observe} loop.
Given the query, the LLM planner first selects the next symbolic action from the bounded action space. The executor then applies this action to the scene KG and returns a grounded observation, which is written back into the interaction context to guide subsequent planning. This process repeats until the reasoning chain reaches a terminal answer.
The loop naturally adapts to question complexity. For simple queries, a single grounded action may be sufficient to produce the answer. For compositional queries involving multiple entities, spatial relations, and status comparisons, the loop unfolds into a multi-step sequence that progressively resolves references, restricts candidate subsets, composes intermediate results, and issues typed queries over the grounded scene state. In this way, the full reasoning program is constructed online through successive grounded executions rather than produced as unconstrained text.
Because each step must instantiate a valid operator and consume verified intermediate results, the resulting reasoning process forms an explicit symbolic trace over the scene KG.
We next describe the agent action space underlying this iterative reasoning process.

%%%%%%%%%%
\begin{figure*}[t]
\centering
\begin{tcolorbox}[
    title=\textbf{Structured Prompt for our Scene Reasoning Agent},
    colback=gray!5,
    colframe=black!60,
    fonttitle=\bfseries,
    sharp corners,
    boxrule=0.8pt,
    left=3mm, right=3mm, top=2mm, bottom=2mm
]
\scriptsize
\begin{itemize}[leftmargin=*, nosep, itemsep=3pt]

    \item \textbf{[Role]} 
    You are an agent that answers fine-grained questions about a 3D driving scene. 
    You must decompose each question into executable operations over the scene, and you must not introduce entities or relations that are not supported by the operators.

    \item \textbf{[Scene-Query Algebra]} 
    The scene is queried through a constrained set-based algebra. 
    \texttt{Resolve(...)} returns the entity set satisfying the specified attribute predicates. 
    \texttt{RelSelect(ref, rel, ...)} returns the entities satisfying the specified predicates and relation \texttt{rel} with respect to reference object \texttt{ref}. 
    \texttt{Intersect(A,B)} composes two entity sets. 
    \texttt{Count(V)} and \texttt{Exists(V)} query the size and non-emptiness of an entity set. 
    \texttt{GetType(V)} and \texttt{GetStatus(V)} return a canonical type or status label from \texttt{V}. 
    \texttt{SameStatus(A,B)} returns \texttt{yes} iff the two entity sets contain a valid pair with the same status.

\item \textbf{[Reasoning Rules]} 
All intermediate variables must be entity sets, except typed outputs from \texttt{Count}, \texttt{Exists}, \texttt{GetType}, \texttt{GetStatus}, and \texttt{SameStatus}. 
Spatial relations follow the ego-centric vocabulary $\mathcal{R}$. 
Reference sets are reduced to their canonical representative, and single-object relational references use the nearest valid instance. 
Arguments and outputs must stay within the bounded scene schema, including categories $\mathcal{C}$, statuses $\mathcal{S}$, relations $\mathcal{R}$, binary answers, and integers.

    \item \textbf{[Exemplars]}
    \begin{itemize}[leftmargin=1.5em, itemsep=1pt]
        %\item \textbf{Q1 (Existence)} \textit{``Are there any parked trucks?''} \\
       % \textbf{Program:} \texttt{trucks = Resolve(type='truck', status='parked'); Exists(trucks)}

        %\item \textbf{Q2 (Relational Counting)} \textit{``How many cars are in front of the standing pedestrian?''} \\
        %\textbf{Program:} \texttt{ped = Resolve(type='pedestrian', status='standing');} \\
        %\phantom{\textbf{Program:}} \texttt{cars = RelSelect(ped, 'front', type='car');} \\
        %\phantom{\textbf{Program:}} \texttt{Count(cars)}
        \item \textbf{Question:} \textit{``Does the car to the left of the stopped truck have the same status as the bus in front of the cyclist?''} \\
        \textbf{Action:}  \texttt{ truck = Resolve(type='truck', status='stopped');} \\
        \phantom{\textbf{Program:}} \texttt{car = RelSelect(truck, 'left', type='car');} \texttt{cyclist = Resolve(type='cyclist');} \\
        \phantom{\textbf{Program:}} \texttt{bus = RelSelect(cyclist, 'front', type='bus');} \texttt{SameStatus(car, bus)}
    \end{itemize}

    \item \textbf{[New Task]} 
    \textbf{Question:} \{QUESTION\}

\end{itemize}
\end{tcolorbox}
\vspace{-0.9em}
\caption{Structured system prompt for the LLM planner.}
\label{fig:tool-prompt}
\vspace{-1em} 
\end{figure*}
%%%%%%%%%%%%

\subsection{LLM Agent Action Space}
\label{subsec:atomic-ops}

We define the agent action space as a bounded scene-query algebra over the scene KG. Its central design principle is that every intermediate state must remain grounded in entities supported by the refined scene representation, rather than latent objects introduced by the language model. Formally, let $\mathcal{V}_\tau$ denote the entity nodes in the scene KG at frame $\tau$, and represent an intermediate reasoning state as an entity subset $\mathcal{U}\subseteq\mathcal{V}_\tau$. Each node in $\mathcal{V}_\tau$ carries a category label in $\mathcal{C}$ and a status label in $\mathcal{S}$.
When a subsequent operation requires a unique reference object or attribute query, we map $\mathcal{U}$ to a deterministic representative $\operatorname{rep}(\mathcal{U})$, defined by the current canonical ordering of the subset.

%We define the agent action space as bounded tools over the scene KG. Its central design principle is that every intermediate state must remain grounded in entities supported by the refined scene representation, rather than latent objects introduced by the language model. Formally, let $\mathcal{V}_\tau$ denote the entity nodes in the scene KG at frame $\tau$. Each node $v_i\in\mathcal{V}_\tau$ is associated with an object category $c_i\in\mathcal{C}$, a status label $s_i\in\mathcal{S}$, and a ground-plane position $\mathbf{p}_i\in\mathbb{R}^2$, where $\mathcal{C}$ and $\mathcal{S}$ denote the finite vocabularies of categories and statuses, respectively. We represent an intermediate reasoning state as an entity subset $\mathcal{U}\subseteq\mathcal{V}_\tau$.When a subsequent operation requires a unique reference object or attribute query, we map $\mathcal{U}$ to a deterministic representative $\operatorname{rep}(\mathcal{U})$, defined by the current canonical ordering of the subset.

\textbf{Entity Resolution.}
The first primitive maps a linguistic object description into a grounded entity subset. Let $\pi$ denote a Boolean predicate over node attributes, for example $c_i=\texttt{car}\wedge s_i=\texttt{moving}$. We define $\textsf{Resolve}(\pi)$=$\{\, v_i \in \mathcal{V}_\tau \mid \pi(v_i)\,\}$.
Thus, \textsf{Resolve} converts a natural-language description into an executable grounded state over perceived scene entities.

\textbf{Relational Restriction.}
To support spatially conditioned queries, we define an ego-centric relational selection operator. Let $\mathbf{d}_{\mathrm{ego}}\in\mathbb{R}^2$ denote the ego heading vector, and let $\mathcal{R}$ be a finite directional vocabulary. Given a reference subset $\mathcal{U}_{\mathrm{ref}}$ and a candidate predicate $\pi$, we first obtain the anchor node $v_{\mathrm{ref}}=\operatorname{rep}(\mathcal{U}_{\mathrm{ref}})$ with ground-plane position $\mathbf{p}_{\mathrm{ref}}$. For each candidate node $v_j$ with position $\mathbf{p}_j$, we compute the signed ego-centric angle $\theta_{\mathrm{ref},j}=\angle_{\mathrm{signed}}(\mathbf{p}_j-\mathbf{p}_{\mathrm{ref}}, \mathbf{d}_{\mathrm{ego}})\in(-\pi,\pi]$.
We then quantize $\theta_{\mathrm{ref},j}$ into a directional relation $r\in\mathcal{R}$ through a deterministic mapping $\Phi_{\mathrm{angle}}:(-\pi,\pi]\rightarrow\mathcal{R}$. 
In practice, $\mathcal{R}=\{\texttt{front}$, $\texttt{front\_left}$, $\texttt{back\_left}$, $\texttt{back}$, $\texttt{back\_right}$, $\texttt{front\_right}\}$, which aligns language-level directional expressions with ego-centric geometry. 
Accordingly, we define
\begin{equation*}
\mathcal{U}_{r}
=
\{\, v_j\in\mathcal{V}_\tau \mid \pi(v_j)\wedge \Phi_{\mathrm{angle}}(\theta_{\mathrm{ref},j})=r \,\}.
\end{equation*}
\begin{equation*}
\textsf{RelSelect}(\mathcal{U}_{\mathrm{ref}}, r, \pi)
=
\operatorname{Sort}_{d(\cdot,v_{\mathrm{ref}})}(\mathcal{U}_{r}).
\end{equation*}
where the result is returned in nearest-first order. In this way, \textsf{RelSelect} grounds spatial language in executable geometry while providing a deterministic ordering for downstream reference selection.

\textbf{Set Composition and Typed Queries.}
Compositional questions are expressed through set composition and typed queries over grounded subsets. We define the basic set-level operators as
\begin{equation*}
\begin{aligned}
    \textsf{Intersect}(\mathcal{U}_A,\mathcal{U}_B) &= \mathcal{U}_A \cap \mathcal{U}_B,\\
    \textsf{Count}(\mathcal{U}) &= |\mathcal{U}|,\\
    \textsf{Exists}(\mathcal{U}) &\in \{\texttt{yes},\texttt{no}\},
\end{aligned}
\label{eq:set_ops}
\end{equation*}
where \textsf{Exists} returns \texttt{yes} iff $\mathcal{U}\neq\emptyset$. We further define typed queries over selected subsets:
\begin{equation*}
\begin{aligned}
    \textsf{GetType}(\mathcal{U}) &= c\!\left(\operatorname{rep}(\mathcal{U})\right)\in\mathcal{C},\\
    \textsf{GetStatus}(\mathcal{U}) &= s\!\left(\operatorname{rep}(\mathcal{U})\right)\in\mathcal{S},\\
    \textsf{SameStatus}(\mathcal{U}_A,\mathcal{U}_B) &\in \{\texttt{yes},\texttt{no}\},
\end{aligned}
\end{equation*}
where $c(\cdot)$ and $s(\cdot)$ return the category and status label of the selected node, respectively, and \textsf{SameStatus} returns \texttt{yes} iff
$s(\operatorname{rep}(\mathcal{U}_A)) = s(\operatorname{rep}(\mathcal{U}_B))$.
Together, \textsf{Resolve}, \textsf{RelSelect}, and the operators above form a compact and executable scene-query algebra for fine-grained reasoning over the scene KG. For example, the question \emph{``How many cars are in front of the standing pedestrian?''} can be expressed as
$\textsf{Count}(\textsf{RelSelect}(\textsf{Resolve}(\pi_{\mathrm{ped,stand}}), \texttt{front}, \pi_{\mathrm{car}}))$,
where $\pi_{\mathrm{ped,stand}}$ selects standing pedestrians and $\pi_{\mathrm{car}}$ selects cars.

\subsection{LLM Planner with In-Context Learning}
\label{subsec:tool-planning}
Given a natural-language query $q$, {\systemname} uses the LLM planner to incrementally construct an executable action sequence $\Pi(q)=(\mathrm{op}_1,\mathrm{op}_2,\dots,\mathrm{op}_K)$ over the bounded action space defined above. Rather than generating the full program in a single pass, the planner operates within the Plan--Execute--Observe loop, where each step is determined jointly by the user query, the structured prompt, and the grounded interaction context accumulated so far. In this way, the LLM is responsible for analyzing the query and selecting valid operator calls over grounded intermediate states.
However, collecting large-scale supervised fine-tuning data for question-driven tool invocation is prohibitively expensive, as it requires substantial manual annotation. Therefore, instead of relying on task-specific fine-tuning, we adopt an in-context learning paradigm to induce planning behavior from a pretrained LLM.

Specifically, we construct a structured system prompt $\mathcal{P}$, shown in Figure~\ref{fig:tool-prompt}. The prompt first assigns the model a planner role for fine-grained 3D scene question answering, and then explicitly specifies the scene-query algebra as its admissible action space, including the available operators and their semantics. It further encodes the execution rules required for valid reasoning, such as maintaining intermediate variables as grounded entity subsets, reducing reference sets to deterministic representatives when needed, interpreting spatial relations in the ego frame, and constraining all arguments and outputs to the bounded schema $(\mathcal{C},\mathcal{S},\mathcal{R})$. Finally, we provide in-context Question--Program exemplars that demonstrate how representative queries should be decomposed into executable operator sequences. As a result, the prompt enables grounded query-to-program planning under explicit structural and geometric constraints, thereby completing the KG-based reasoning for fine-grained scene question answering.

\begin{table*}[t]
  \centering
  \renewcommand{\arraystretch}{1.1}
  \scriptsize
  \resizebox{0.7\textwidth}{!}{%
  \begin{tabular}{l *{9}{c}}
    \toprule
    \multirow{2}{*}{Models} & 
      \multicolumn{6}{c}{NuScenes-QA~\cite{qian2024nuscenes}} &
      \multicolumn{3}{c}{GVQA~\cite{sima2024drivelm}} \\
    \cmidrule(lr){2-7}\cmidrule(lr){8-10}
      & \multicolumn{1}{c}{Acc $\uparrow$ (Exist)}
      & Acc $\uparrow$ (Count)
      & Acc $\uparrow$ (Object)
      & Acc $\uparrow$ (Status)
      & Acc $\uparrow$ (Comparison)
      & Overall Acc $\uparrow$
      & \multicolumn{1}{c}{SPICE $\uparrow$}
      & METEOR $\uparrow$
      & CIDEr $\uparrow$ \\
    \midrule
    \rowcolor{gray!15} FocalFormer3D$^{\dagger}$~\cite{chen2023focalformer3d} & \multicolumn{1}{|c}{50.28} & 0.01 & 0.55 & 0.22 & 47.36 & 24.02 & \multicolumn{1}{|c}{14.41} & 0.21 & 0.56 \\
    RayDN$^{\dagger}$~\cite{liu2024ray} & \multicolumn{1}{|c}{44.13} & 0.04 & 1.48 & 0.28 & 44.38 & 21.46 & \multicolumn{1}{|c}{11.58} & 0.20 & 0.46 \\
    \rowcolor{gray!15} IS-Fusion$^{\dagger}$~\cite{yin2024fusion} & \multicolumn{1}{|c}{52.70} & 0.04 & 0.42 & 0.67 & 44.87 & 24.86 & \multicolumn{1}{|c}{11.20} & 0.19 & 0.44 \\
    \midrule
    LLaVA-v1.6-mistral-7$^{\ddagger}$ & \multicolumn{1}{|c}{48.64} & 6.83 & 1.48 & 5.09 & 52.80 & 26.17 & \multicolumn{1}{|c}{43.75} & 0.11 & 0.38 \\
    \rowcolor{gray!15} Internvl-3.5-8b$^{\ddagger}$ & \multicolumn{1}{|c}{49.86} & 11.16 & 35.74 & 22.29 & 38.72 & 34.28 & \multicolumn{1}{|c}{32.66} & 0.13 & 0.34 \\
    Qwen3Vision-8B$^{\ddagger}$ & \multicolumn{1}{|c}{64.16} & 10.30 & 27.66 & 26.11 & 39.63 & 39.53 & \multicolumn{1}{|c}{36.44} & 0.22 & 0.50 \\
    \midrule
    \rowcolor{gray!15} DriveLM$^{\dagger}$~\cite{sima2024drivelm}  & \multicolumn{1}{|c}{77.29} & 16.92 & 53.67 & 50.64 & 67.41 & 56.42 & \multicolumn{1}{|c}{42.16} & 0.42 & 3.53 \\
    LiDAR-LLM$^{\dagger}$~\cite{yang2025lidar} & \multicolumn{1}{|c}{76.41} & 15.84 & 50.63 & 48.00 & 69.11 & 55.08 & \multicolumn{1}{|c}{41.09} & 0.42 & 3.35 \\
    \rowcolor{gray!15} MAPLM$^{\dagger}$~\cite{cao2024maplm} & \multicolumn{1}{|c}{79.29} & 18.45 & 60.67 & 58.38 & 70.55 & 60.17 & \multicolumn{1}{|c}{37.12} & 0.36 & 3.09 \\
    CREMA$^{\dagger}$~\cite{yu2025crema}  & \multicolumn{1}{|c}{73.17} & 15.69 & 53.69 & 52.63 & 72.03 & 55.26 & \multicolumn{1}{|c}{39.05} & 0.39 & 3.27 \\
    \midrule
    \rowcolor{gray!15} \systemname-KG$^{\ddagger}$ (Ours) & \multicolumn{1}{|c}{74.67} & 64.46 & 54.30 & 57.84 & 53.23 & 65.04 & \multicolumn{1}{|c}{42.45} & 0.41 & 3.48 \\
    \midrule
    \systemname-GT$^{\ddagger}$ (Ours) & \multicolumn{1}{|c}{\textbf{88.55}} & \textbf{78.92} & \textbf{80.11} & \textbf{85.72} & \textbf{82.01} & \textbf{84.49} & \multicolumn{1}{|c}{\textbf{43.89}} &\textbf{ 0.43} & \textbf{3.88} \\
    \bottomrule
  \end{tabular}}

\caption{Comparison for \systemname~with baselines on NuScenes-QA~\cite{qian2024nuscenes} and GVQA~\cite{sima2024drivelm}.
Bold numbers indicate the best performance. 
Methods marked with $^{\dagger}$ trained since they have task-specific parameters, whereas methods marked with $^{\ddagger}$ use frozen LLM with few-shot ICL only.}\label{tab:overall}
  \vspace{-0.6cm}
\end{table*}

%\underline{underlined} numbers denote the second-best results.

%We evaluate {\systemname} on NuScenes-QA~\cite{qian2024nuscenes}, a large-scale QA benchmark built on the nuScenes autonomous driving dataset~\cite{caesar2020nuscenes}, which provides synchronized multi-view camera and LiDAR streams in urban traffic scenes. NuScenes-QA augments these multi-modal sequences with 459{,}941 question–answer pairs over 34{,}149 scenes, following the official split of 28{,}130 scenes (376{,}604 questions) for training and 6{,}019 scenes (83{,}337 questions) for testing. 
%\section{Results}

\section{Experimental Setup}
%\subsubsection{Dataset and Evaluation Metrics}\label{subsec:dataset}
\subsection{Datasets}
We evaluate {\systemname} on two large-scale public benchmarks, namely NuScenes-QA~\cite{qian2024nuscenes} and GVQA~\cite{sima2024drivelm}. These two datasets systematically evaluate the fine-grained scene understanding and factual reasoning capabilities of {\systemname} from two levels: structured short-answer and sentence level comprehensive QA.
\textit{(1) NuScenes-QA}~\cite{qian2024nuscenes} is currently the largest public benchmark for factual reasoning in driving scenes. We select 39,704 challenging fine-grained factual reasoning questions, covering five high-level categories: \textit{existence}, \textit{counting}, \textit{object query}, \textit{status query}, and \textit{comparison}. These questions require short structured answers, including binary yes/no responses, integer counts, and categorical labels such as object identity and motion status.
\textit{(2) GVQA}~\cite{sima2024drivelm}  is a QA benchmark for driving understanding with a richer coverage of dimensions. Besides factual understanding, it also includes speed prediction and driving behavior QA. Therefore, we select 104,030 QA tasks oriented towards understanding basic facts.
These questions involve compositional queries over object identity, motion status, existence, and quantity, and their answers are given in sentence form.

%\subsubsection{Evaluation Metrics}\label{subsec:metrics}
\subsection{Metrics}
We use dataset-specific metrics to account for the different answer formats in the two benchmarks, and additionally report inference latency to measure efficiency.
\textbf{(1) Exact Match Accuracy.}
For NuScenes-QA, where answers are short structured outputs, we follow the standard evaluation protocol~\cite{qian2024nuscenes} and report strict exact-match accuracy.
We present accuracy for each question category as well as the overall accuracy.
\textbf{(2) Sentence-level QA Metrics.}
For GVQA~\cite{sima2024drivelm}, where answers are primarily sentence-form factual responses.
We therefore adopt GVQA recommended metrics, including SPICE, METEOR, and CIDEr. 
These are widely used metrics for sentence-level QA evaluation, providing semantic correctness metrics of different dimensions for the generation of factual answers.
We report these metrics using their standard scales, where higher values indicate better performance.
\textbf{(3) Reasoning Time (s).}
To evaluate reasoning efficiency, we measure the average per-question inference latency.
This metric reflects the end-to-end time required to generate an answer for a given question.

%(i) traditional 3D perception SOTA pipelines with a learned QA head, (ii) driving-oriented LLM frameworks that adapt foundation models to the autonomous driving domain, and (iii) frozen foundation VLMs evaluated in a few-shot in-context learning setting.
%\subsubsection{Baseline Methods and \systemname~variants}

\subsection{Baselines}
We compare {\systemname} with three representative groups of baselines: (1) Perception-based pipelines with learned QA heads, (2) Foundation VLMs under few-shot in-context learning, and (3) driving-specialized multimodal LLM/VLM frameworks.
Methods marked with $^{\dagger}$ require task-specific training or fine-tuning, whereas methods marked with $^{\ddagger}$ keep the backbone frozen and use few-shot in-context learning only.

\textbf{(1) Perception-based baselines.}
We consider three SOTA perception backbones: RayDN$^{\dagger}$~\cite{liu2024ray}, FocalFormer3D$^{\dagger}$~\cite{chen2023focalformer3d}, and IS-Fusion$^{\dagger}$~\cite{yin2024fusion}, corresponding to camera-only, LiDAR-only, and camera--LiDAR fusion settings.
We attach benchmark-specific QA heads to their perception outputs.
For NuScenes-QA, which requires short structured answers, we follow the standard setting and pair each detector with the MCAN QA head~\cite{yu2019deep}, which is trained together with the QA task.
For DriveLM, whose answers are sentence-form factual responses, we pair each detector with a BERT-based answer head to produce sentence-level outputs.
These baselines therefore test how far detector outputs can support fine-grained reasoning and QA without explicit knowledge graph.

\textbf{(2) Foundation VLMs.}
To assess the ability of foundation VLMs models to reason and understand fine-grained autonomous driving problems, we evaluated several representative VLMs, including LLaVA-v1.6-Mistral-7B$^{\ddagger}$~\cite{Liu_2024_CVPR}, InternVL-3.5-8B$^{\ddagger}$~\cite{wang2025internvl3_5}, and Qwen3Vision-8B$^{\ddagger}$~\cite{qwen3technicalreport}.
These models take rendered multi-view camera images and the question as input, and directly generate answers in an end-to-end manner.
All model parameters are kept frozen, and adaptation is performed purely through few-shot in-context exemplars.

\textbf{(3) Driving-specialized LLM frameworks.}
We further benchmark against representative recent SOTA baselines for multi-modal driving-scene understanding, including DriveLM $^{\dagger}$~\cite{sima2024drivelm}, LiDAR-LLM$^{\dagger}$~\cite{yang2025lidar}, the official MAP- LM$^{\dagger}$~\cite{cao2024maplm}, and CREMA$^{\dagger}$~\cite{yu2025crema}.
These methods build on foundation LLM backbones and introduce task-specific encoders, projectors, or adapters to better model driving-scene inputs and QA tasks.
They represent strong recent baselines for LLM-based scene understanding in autonomous driving and closely related multimodal reasoning settings.
We follow their official implementations and fine-tune them on both NuScenes-QA~\cite{qian2024nuscenes} and GVQA~\cite{sima2024drivelm} dataset.

\textbf{(4) {\systemname} variants.}
We report two variants of our framework.
{\systemname}-KG is the full system, which constructs the scene KG from predicted perception outputs.
{\systemname}-GT is a variant that builds the KG directly from ground-truth annotations, in order to isolate reasoning performance from perception errors.
Both variants use the same bounded query tool set and the same frozen LLM agent, controlled purely through few-shot in-context learning.

\subsection{Implementation Details}
For both NuScenes-QA~\cite{qian2024nuscenes} and GVQA~\cite{sima2024drivelm} dataset, we randomly split the data into training and test sets with an 8:2 ratio.
%All trainable baselines, including the perception-based pipelines and the driving-specialized LLM frameworks, are trained or fine-tuned on the corresponding training split and evaluated on the est split.
%Foundation VLM baselines use the official released checkpoints and are evaluated in a frozen few-shot in-context learning setting, without any parameter updates. 
%We report the best inference results for all trainable baselines under their corresponding benchmark settings.
We implement {\systemname} using PyTorch and HuggingFace. For {\systemname}, the perception front-end is trained on training dataset, while the reasoning agent uses a frozen Qwen3-7B backbone and is controlled solely through few-shot in-context learning, with no gradient updates.
Experiments are conducted on a workstation equipped with an AMD Ryzen Threadripper 9985WX CPU, 768~GB RAM, and NVIDIA RTX A6000 GPUs.
For edge-side latency evaluation, we use an NVIDIA Jetson Orin NX with 16~GB RAM.
%The platform runs JetPack 6.0 / L4T 36.3.0 on Ubuntu 22.04, and latency is measured under the 10W power mode.

%All experiments are conducted on a unified server equipped with NVIDIA GeForce RTX 3090 GPUs, 16 CPU cores, and 64~GB RAM.

% %%%%%%%%%%%%%%%%%%%%%%%%%%%%%%%%

%  \begin{figure*}[h]
% \centering
% % ----------------
% \begin{minipage}[t]{0.24\textwidth}
%   \centering
%   \includegraphics[height=1.0in]{fig/experiment/coverage_rate.pdf}
%    \vspace{-0.1in}
%   \captionof{figure}{Objects coverage rate by detection backbone.}
%   \label{fig:coverage_rate}
% \end{minipage}
% \hfill
% % ----------------
% \begin{minipage}[t]{0.5\textwidth}  
%   \centering
%   % 1
%   \begin{minipage}[t]{0.49\textwidth}
%     \centering
%     \includegraphics[height=1.0in]{fig/experiment/perception_bythreshould.pdf}
%     %\captionof{subfigure}{...}
%   \end{minipage}%
%   \hfill
%   % 2
%   \begin{minipage}[t]{0.49\textwidth}
%     \centering
%     \includegraphics[height=1.0in]{fig/experiment/perception_ourKG.pdf}
%   \end{minipage}
%   % caption
% \captionsetup{margin={2em,0pt}}
%  \vspace{-0.1in}
%  \captionof{figure}{
%     The perception performance original \\
%     detector (left) and our improved module (right).}
%   \label{fig:comparison}
% \end{minipage}
% \hfill
% \begin{minipage}[t]{0.24\textwidth}
%   \centering
%   \includegraphics[height=1.0in]{fig/experiment/resoning_results.pdf}
%    \vspace{-0.1in}
%   \captionof{figure}{The reasoning performance of \systemname-GT.}
%   \label{fig:resoning_results}
% \end{minipage}
% \end{figure*}
% %%%%%%%%%%%%%%%%%%%%%%%%%%%%%
\begin{figure*}[t]
\centering

% ================= first row =================
\begin{minipage}[t]{0.46\textwidth}
    \centering
    \begin{minipage}[t]{0.46\textwidth}
        \centering
        \includegraphics[width=\linewidth]{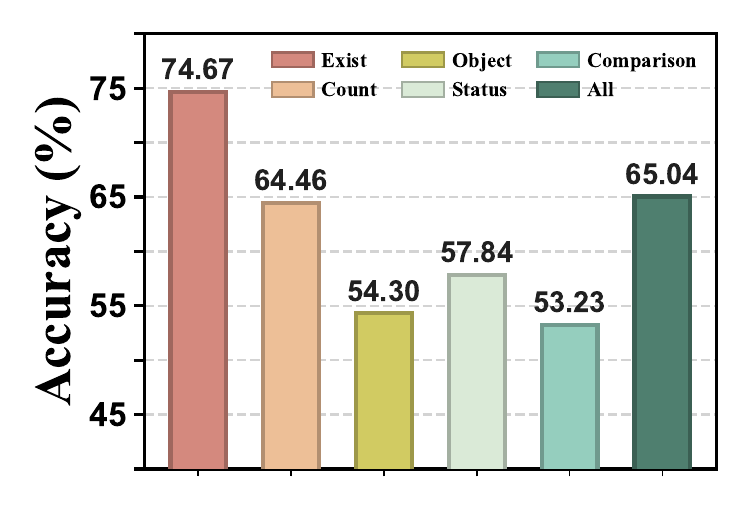}
    \end{minipage}\hfill
    \begin{minipage}[t]{0.46\textwidth}
        \centering
        \includegraphics[width=\linewidth]{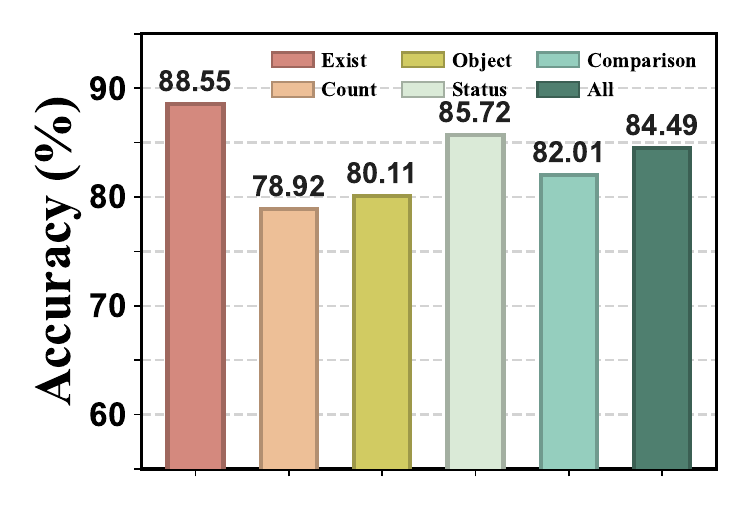}
    \end{minipage}
\vspace{-0.3cm}
    \captionof{figure}{Comparison of {\systemname}-KG (left) and {\systemname}-GT (right) accuracy on different categories.}
    \label{fig:overall_results_cat}
\end{minipage}\hfill
\begin{minipage}[t]{0.48\textwidth}
    \centering
    \begin{minipage}[t]{0.48\textwidth}
        \centering
        \includegraphics[width=\linewidth]{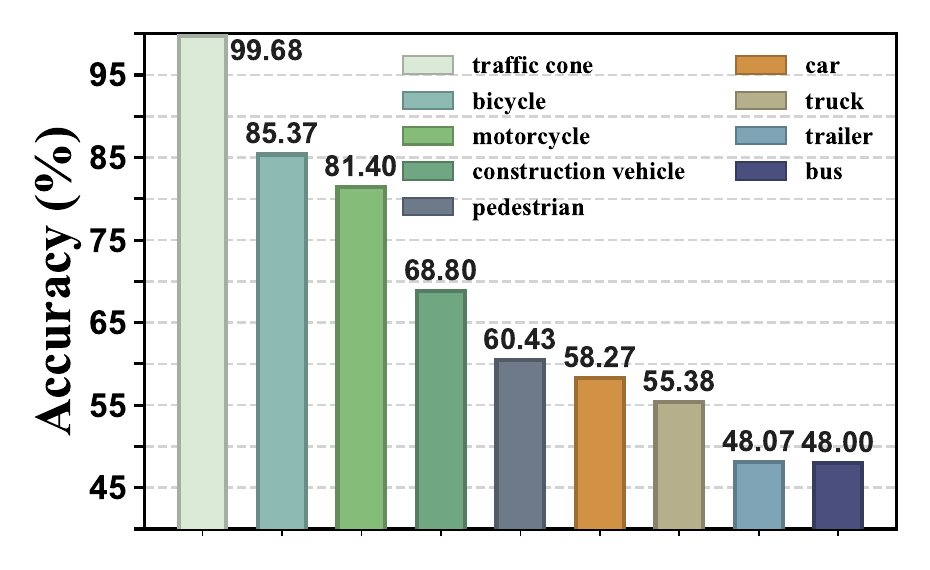}
    \end{minipage}\hfill
    \begin{minipage}[t]{0.48\textwidth}
        \centering
        \includegraphics[width=\linewidth]{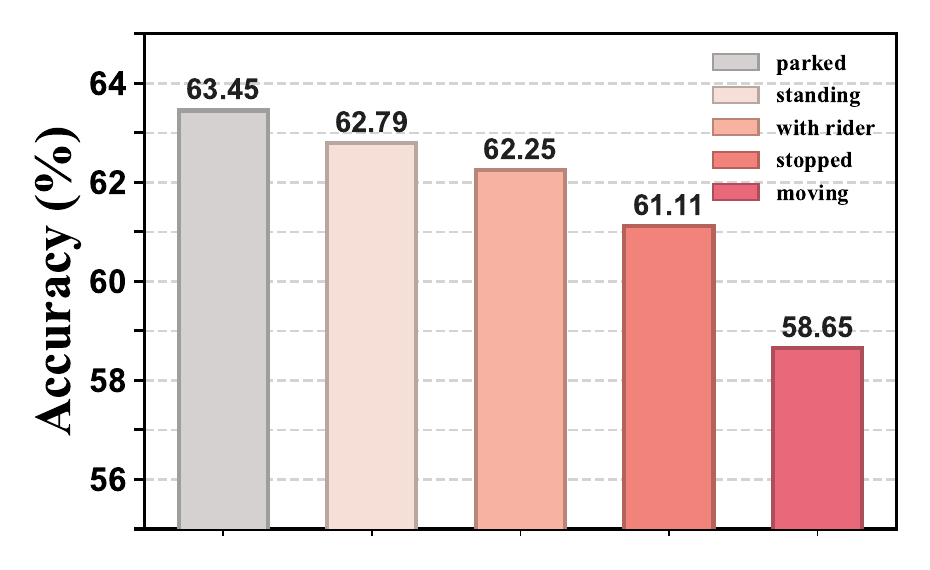}
    \end{minipage}
\vspace{-0.3cm}
    \captionof{figure}{Comparison of {\systemname}-KG accuracy across object categories (left) and attributes (right).}
    \label{fig:overall_results_objatt}
\end{minipage}

\vspace{0.23cm}

% ================= second row =================
\begin{minipage}[t]{0.26\textwidth}
    \centering
    \includegraphics[width=\linewidth]{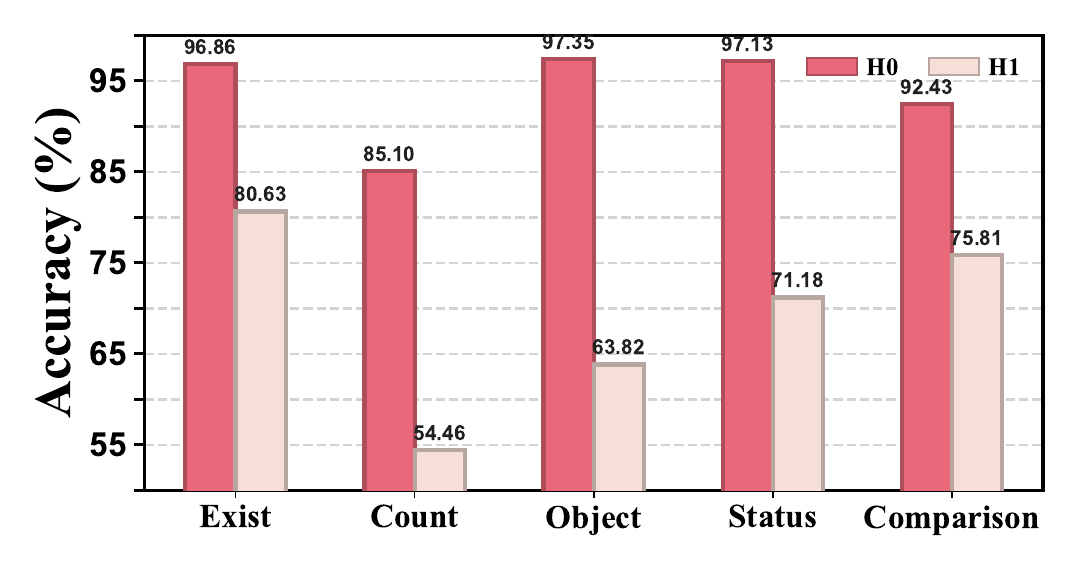}
    \vspace{-0.6cm}
    \captionof{figure}{{\systemname}-GT Accuracy for reasoning difficulty.}
    \label{fig:overall_results_hop}
\end{minipage}\hfill
\begin{minipage}[t]{0.20\textwidth}
    \centering
    \includegraphics[width=\linewidth]{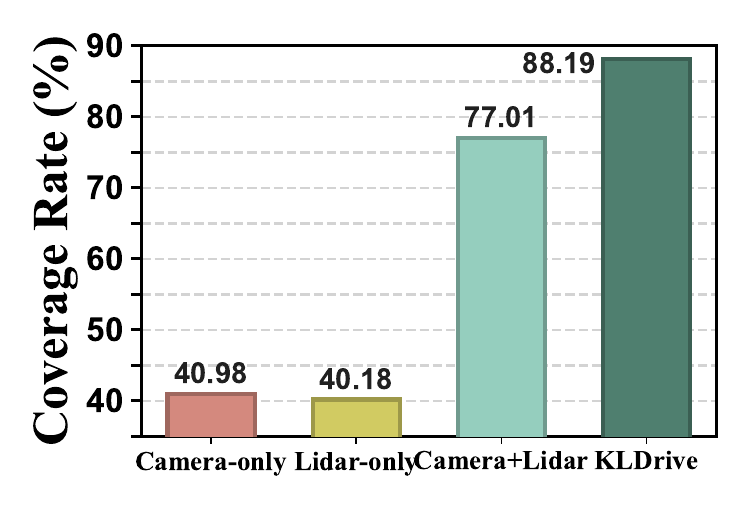}
    \vspace{-0.6cm}
    \captionof{figure}{Coverage of each evidence space.}
    \label{fig:overall_results_cov}
\end{minipage}\hfill
\begin{minipage}[t]{0.22\textwidth}
    \centering
    \includegraphics[width=\linewidth]{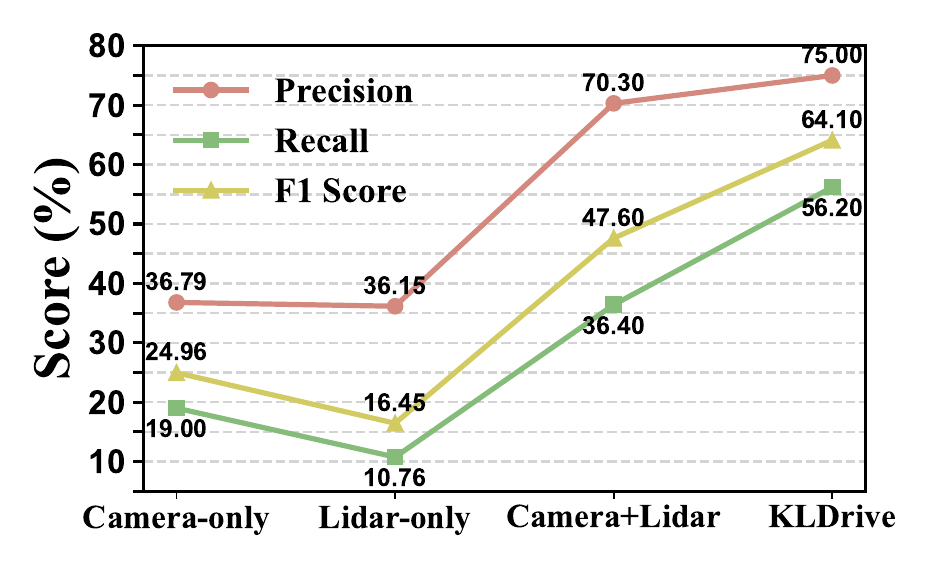}
    \vspace{-0.6cm}
    \captionof{figure}{Effect of EBM-based refinement model.}
    \label{fig:overall_results_ebm}
\end{minipage}\hfill
\begin{minipage}[t]{0.22\textwidth}
    \centering
    \includegraphics[width=\linewidth]{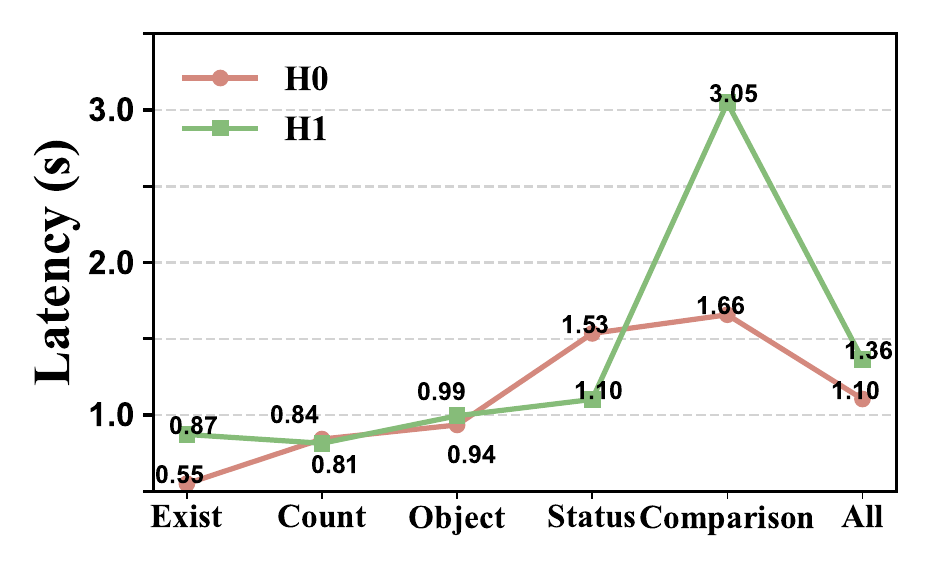}
    \vspace{-0.6cm}
    \captionof{figure}{Latency by QA category and difficulty.}
    \label{fig:overall_results_latency}
\end{minipage}
\vspace{-0.3cm}
\end{figure*}

\section{Experimental Results}
\subsection{Overall Results}
\label{subsec:overall}

Table~\ref{tab:overall} summarizes the overall performance of all methods on NuScenes-QA~\cite{qian2024nuscenes} and GVQA~\cite{sima2024drivelm} dataset.
Across both benchmarks, the three baseline families exhibit a clear results: perception-based methods perform worst, and general foundation VLMs are improved, while driving-specialized LLM frameworks provide the strongest competing results after task-specific fine-tuning.
Compared with these baselines, {\systemname} further improves fine-grained scene understanding and reasoning.
On NuScenes-QA, {\systemname}-KG achieves the best overall accuracy of 65.04\%, outperforming the strongest baseline, MAPLM, by 4.87 points.
When the perceived result is completely accurate, the reasoning performance further increases to 84.49\% with {\systemname}-GT.
On GVQA, {\systemname}-KG also remains highly competitive against the strongest task-specific fine-tuned driving-oriented LLM/ MVLM frameworks, achieving the best SPICE score of 42.45 while maintaining comparable METEOR and CIDEr.
Moreover, {\systemname}-GT attains the best performance on all three metrics, reaching 43.89 SPICE, 0.43 METEOR, and 3.88 CIDEr.

Beyond the numerical gains, these results highlight the key advantage of the new paradigm introduced by {\systemname}: it enables fact-grounded, transparent, and traceable QA reasoning.
Unlike prior end-to-end methods, which map sensor inputs directly to answers through opaque trained representations, {\systemname} first aggregates multi-sensor evidence from the scene and constructs a scene knowledge graph, and then uses the LLM as a planner to invoke tools step by step within a designed action space.
Because the action space consists of deterministic operations over the scene KG, every answer is grounded in detected scene facts rather than produced by a potentially correct but unverified guess.
The entire reasoning process is transparent, auditable, and traceable.
Through explicit tool calls in the action space and the intermediate results they return, each answer can be traced step by step to concrete objects, attributes, and relations in the evidence space.
As a result, the agent will not produce speculative answers that are not supported by detected evidence.

This advantage is especially evident on \textit{counting} tasks, where {\systemname}-KG reaches 64.46\% accuracy, while the best baseline achieves only 18.45\%, yielding a large gain of 46.01 points. However, the LLM-related baselines achieve an average accuracy of only 13.60\%, suggesting serious hallucination issues.
Compared with conventional perception-driven pipelines, {\systemname} delivers significantly gains across all tasks.
Notably, these perception methods also serve as the detection backbones of {\systemname}, therefore further highlights the necessity of our reasoning module design.
Although such perception-driven methods can detect scene objects, they do not explicitly organize these detections into a structured scene knowledge graph or perform effective understanding over them.
Overall, by transforming detected evidence into an executable scene KG and coupling it with a tool-driven reasoning agent, {\systemname} effectively bridges perception and reasoning, achieving the best overall performance compared with the latest baselines.

\subsection{Analysis of Reasoning Tasks}
We further analyze performance across QA categories and reasoning difficulty levels to better understand the performance of {\systemname}.

\textbf{QA Categories.} Figure~\ref{fig:overall_results_cat} shows that both {\systemname}-KG and {\systemname}-GT perform best on \textit{existence} questions, which is consistent with intuition since existence queries are the simplest.
For {\systemname}-KG, \textit{object} and \textit{comparison} are the most challenging categories.
However, once scene facts are replaced with ground-truth annotations, these two categories improve substantially, with gains of 25.81 and 28.78 points, respectively.
This suggests that fine-grained semantic QA is highly sensitive to the fidelity of entities, attributes, and relations in the scene representation, and that improving perception quality remains critical for further end-to-end gains.
In contrast, \textit{counting} shows a relatively smaller improvement and remains one of the most difficult tasks even under ground-truth scene facts, suggesting that its difficulty lies more in the execution of complex reasoning than merely in the quality of basic scene facts.

\textbf{Objects and Attributes.}
We further analyze the {\systemname}-KG results from the perspectives of object categories and attributes, as shown in Figures~\ref{fig:overall_results_objatt}.
Overall, attribute-related questions are more accurate and more stable than object-related ones: the weighted accuracy is 61.50\% for attribute reasoning, compared with 55.29\% for object reasoning.
Moreover, the variation across attributes is relatively small, whereas object-related performance differs much more substantially across categories.
Specifically, objects with more distinctive visual characteristics and clearer category boundaries, such as traffic cones and bicycles, are easier to recognize, and the corresponding QA accuracy is also higher; in particular, traffic-cone-related questions are nearly saturated at 99.6\%.
In contrast, categories with more similar appearance and semantics, such as trucks and buses, remain considerably more challenging and yield lower QA accuracy.
 
\textbf{Difficulty Levels.} Next, we analyze the performance of {\systemname} under different reasoning difficulty levels. To exclude the effect of perception noise, we study the results of {\systemname}-GT, as shown in Figure~\ref{fig:overall_results_hop}.
Here, \textit{H0 (zero-hop)} denotes factual queries that can be answered directly from the attributes of a single object, e.g., ``Is there a truck in the scene?'', whereas \textit{H1 (one-hop)} denotes compositional queries that require relational or spatial reasoning across multiple entities, e.g., comparing the status of two vehicles at different locations.
The results show that H0 achieves consistently high performance across all categories, while H1 is substantially more challenging. This gap is especially pronounced for \textit{counting}, where the accuracy drops from 85.10 (H0) to 54.46 (H1). These findings indicate that once scene facts are accurately established, the remaining reasoning bottleneck mainly lies in multi-entity compositional reasoning, particularly multi-object counting and reasoning under complex spatial relations.

%%%%%%%%%%%%%%%%%%%%%%%%%
\begin{table}[h]
\centering
\resizebox{0.48\textwidth}{!}{\begin{tabular}{lcccccc}
\toprule
 & Exist & Count & Object & Status & Comparison & Overall Acc \\
\midrule
w/o KG Generation Module      & 53.39 &  12.5 &  16.54 & 8.47 & 48.85 & 28.10 \\
w/ KG Module (\systemname-KG)& 74.67 & 64.46 & 54.30 & 57.84 & 53.23 & 65.04 \\
\midrule
w/o LLM Agent Module            & 59.54 & 20.71 & 71.14 & 62.31 & 55.52 & 53.50 \\
w/ LLM Agent Module (\systemname-GT) & 85.55 & 78.92 & 80.11 & 85.72 & 82.01&84.49 \\
\bottomrule
\end{tabular}}
 %\vspace{0.1in}
\caption{Ablation comparison (accuracy\%) of each functional module in \systemname\, illustrating the impact of KG generation model and designed LLM agent.}
\label{tab:ablation}
\vspace{-0.5cm}
\end{table}

%%%%%%%%%%%%%%%%%%%%%%%%%

\subsection{Ablation Study}
\label{subsec:AblationStudies}
To quantify the contribution of each component in \systemname, we perform ablation study on the KG Generation model and the designed LLM agent, as summarized in Table~\ref{tab:ablation}.

\textbf{Perception Model.}
We first validate the necessity and effectiveness of multi-source evidence aggregation and EBM-based evidence refinement.
For fine-grained and complex reasoning in 3D scenes, relying on a single-modality detector is insufficient to provide a sufficiently complete fact space. To quantify this, we match detector outputs from different modalities with ground-truth objects and compute the \emph{coverage rate}, defined as the fraction of ground-truth objects covered by at least one detection. As shown in Figure~\ref{fig:overall_results_cov}, the coverage rates of the camera-only (RayDN~\cite{liu2024ray}) and LiDAR-only (FocalFormer3D~\cite{chen2023focalformer3d}) detectors are only 40.98\% and 40.18\%, respectively, while the camera--LiDAR fusion detector IS-Fusion~\cite{yin2024fusion} increases the coverage to 77.01\%. On top of this, by further aggregating multi-source detection evidence, we improve the coverage rate to 88.19\%.

However, higher coverage alone is still insufficient for reliable reasoning, because the fused candidates still contain substantial noise, duplicates, and attribute conflicts. To address this, we further introduce EBM-based evidence refinement, which performs consistency modeling and global selection over multi-source candidates. As shown in Figure~\ref{fig:overall_results_ebm}, compared with directly using the raw fused candidates, {\systemname} significantly improves the precision, recall, and F1-score of detected objects. This shows that EBM refinement can not only effectively recover missed objects, but also mitigate conflicts among multi-source candidates, thereby producing a more reliable structured scene representation for downstream reasoning.

We then study the gain brought by the EBM-based KG construction method. If we directly aggregate the outputs of multi-source detectors to construct a KG and perform reasoning, the overall accuracy is only 28.10\%. In contrast, {\systemname}-KG leverages temporal association and interpolation to recover missing objects, and uses energy-based evidence refinement to select more consistent and reliable entities and attributes from multi-source candidates, improving the overall accuracy to 65.04\%, a gain of 36.94 points. This improvement is particularly evident on several key categories, including \textit{counting}, \textit{object}, and \textit{status}.
Overall, the ablation results show that raw detection outputs alone are insufficient to support reliable reasoning. Through EBM-based evidence refinement and structured KG construction, {\systemname} can more effectively recover missed objects, alleviate attribute conflicts, and build a fact space that is better suited for downstream compositional reasoning.

\textbf{LLM Agent.}
We further evaluate the contribution of our LLM-agent design. Specifically, we feed the constructed KG into the same Qwen3-7B backbone for end-to-end reasoning and answer generation. The overall accuracy drops to 44.38\%, showing that for complex fine-grained reasoning tasks, LLMs are prone to hallucination without an explicitly defined action space and tools. This issue is most evident on \textit{counting} and \textit{comparison}, where the performance gaps reach 58.21 and 26.49 points, respectively, indicating that compositional reasoning such as multi-entity filtering and comparative aggregation particularly depends on explicit tool invocation and step-by-step execution.

We then compare the importance of two key components in the {\systemname} agent: few-shot in-context exemplars and the \emph{Plan--Execute--Observe} loop. Removing the in-context exemplars used to guide the LLM planner to invoke tools reduces the overall accuracy from 84.49\% to 71.54\%. Removing the \emph{Plan--Execute--Observe} loop further reduces it to 62.35\%. This shows that the former is crucial for stably generating correct tool-call sequences, while the latter enables the model to continually correct its reasoning through intermediate observations and thus reduces error accumulation in complex compositional reasoning. Overall, these results show that the core value of the LLM agent lies in transforming fine-grained and complex QA into a constrained, executable, and verifiable reasoning process, and the performance gains are substantial.
%%%%%%%%%%%%%%%%%%%%%%%%

\subsection{Latency Analysis}
We evaluate the reasoning latency of {\systemname}. As shown in Figure~\ref{fig:overall_results_latency}, the system achieves an average latency of 1.26~s per question on an NVIDIA RTX A6000. Among all categories, \textit{Exist} is the fastest, with an average latency of only 0.71~s, whereas comprehensive sentence-level outputs are the slowest, averaging 2.95~s. This is consistent with their task complexity: the former usually requires only direct decisions over a small number of scene facts, whereas the latter involves a longer chain of comprehensive reasoning and sentence-level answer generation.
We also observe latency differences across reasoning difficulty levels, which are especially pronounced for complex \textit{comparison} questions. This is because H1 typically involves multi-entity localization, spatial relation resolution, and multiple tool calls, resulting in longer execution chains and higher reasoning overhead. Overall, these results show that the latency of {\systemname} scales naturally with question complexity, but remains acceptable in exchange for a transparent, traceable, and verifiable reasoning process.

We further evaluate the system on an edge device, NVIDIA Jetson Orin, where the average latency is 57.05~s and follows the same trend as on the A6000. 
{\systemname} is designed for offline fine-grained reasoning over complex driving scenes, rather than for online planning tasks with strict real-time response requirements.
Therefore, considering the highly limited compute capability of Jetson and the complexity of the reasoning task, this latency is acceptable. The result indicates that {\systemname} can be deployed on resource-constrained edge platforms. In future work, we will further explore lighter LLM backbones, model quantization, and reasoning-chain compression to further reduce latency and resource overhead in edge deployment.

\section{Related Work}
%Table~\ref{tab:sota_comparison} summarizes representative SOTA methods and their key properties.As shown, 3D perception pipelines offer strong perception but lack explicit reasoning, while driving-oriented LLM frameworks provide reasoning but also rely on heavy task-specific training and remain opaque. \systemname\ is designed to combine strong perception and reasoning without task-specific finetuning, while keeping the reasoning process explainable.

%Table~\ref{tab:sota_comparison} summarizes representative SOTA methods and their key properties. As shown, 3D perception pipelines provide strong perception but lack explicit reasoning, whereas driving-oriented LLM frameworks offer reasoning but also require heavy task-specific training and remain opaque. \systemname\ combines strong perception with explicit, explainable reasoning without task-specific finetuning.

\subsection{3D Scene Understanding}
3D scene understanding in autonomous driving has been extensively studied along three lines: vision-based BEV methods, LiDAR-based detectors, and multi-modal fusion. 
Vision-based pipelines construct BEV features from multi-view images and perform 3D detection in BEV space~\cite{huang2021bevdet,li2022bevformer,li2024bevnext,liu2024ray}. 
LiDAR-based approaches exploit sparse geometric encodings and center-based paradigms for robust long-range detection~\cite{yin2021center,zhou2022centerformer,chae2024lidar}. 
Fusion frameworks combine camera and LiDAR streams through BEV alignment or instance-level interaction to improve  accuracy~\cite{liang2022bevfusion,jiao2023msmdfusion,yin2024fusion,chen2023focalformer3d}. 
These methods constitute the current SOTA in 3D scene perception, but they are optimized for detection metrics and produce only bounding boxes, without explicitly encoding higher-order spatial relations and supporting fine-grained factual reasoning about the scene.
In this work, {\systemname} transforms perception outputs into a scene knowledge graph that explicitly captures object states and spatial relations for fine-grained factual reasoning.

\subsection{VLM \& Agent for Autonomous Driving}

Recent work adapts large VLMs to driving, aiming to unify perception, prediction, and planning in a language interface. 
DriveGPT4~\cite{xu2024drivegpt4} and DriveVLM~\cite{tian2024drivevlm} jointly model videos and other modalities with natural-language explanations or low-level control commands; 
GPT-Driver\cite{mao2023gpt} and LMDrive\cite{shao2024lmdrive} use LLMs as decision modules, conditioned on multimodal perception features from cameras and LiDAR. 
To better structure multi-step driving decisions, Agent-Driver~\cite{mao2023language} further treats an LLM as a driving agent, combining tool use to execute the planning workflow.
Domain-specific frameworks such as DriveLM, LiDAR-LLM, MAPLM, and CREMA build on general-purpose VLM backbones, attach task-specific encoders for images and point clouds, and fine-tune on large autonomous-driving data to perform language-driven perception and reasoning~\cite{sima2024drivelm,yang2025lidar,cao2024maplm,yu2025crema}. 
These systems represent the current LLM-based SOTA on driving benchmarks, but they map sensor features to textual outputs in an end-to-end manner, making their reasoning opaque and tightly coupled to large-scale task-specific training.
Our empirical study further shows that these models still hallucinate heavily on fine-grained factual queries, particularly for counting and spatial relations.
In contrast, \systemname\ uses a frozen 7B LLM as a tool-using planner over a structured 3D scene KG, deriving answers via explicit tool execution on the KG rather than unconstrained text generation, which reduces hallucinations significantly and obviates additional task-specific finetuning.

%%%%%%%%%%%%%%%

%\vspace{-0.3cm}
\subsection{KG-Augmented LLM Reasoning}
Recent work shows that augmenting LLMs with KG can reduce hallucination and improve traceability~\cite{agrawal2024can,sun2023think,ye2024correcting,ma2025knowledge,nishat2025aligning}.
GraphRAG~\cite{graphrag-msr} and SubgraphRAG~\cite{li2024simple} construct or retrieve subgraphs as structured evidence for long-document and multi-hop textual QA. 
KELP~\cite{liu2024knowledge}, GLAME~\cite{zhang2024knowledge}, and related methods~\cite{coppolillo2025injecting} inject KG paths or embeddings into LLMs to guide factual reasoning and constrain model editing, while \textit{Think-on-Graph}~\cite{sun2023think} and KG-CoT~\cite{zhao2024kg} explicitly treat LLMs as graph search agents that generate reasoning chains over KG to obtain interpretable answers.
These studies demonstrate that combining structured KG with tool or chain-based reasoning can mitigate hallucinations. 
However, they are primarily developed for static symbolic knowledge graphs derived from text or curated knowledge bases, and do not address reasoning over noisy, multi-sensor, spatio-temporal 3D scenes with ego-centric geometry as required in autonomous driving.
\systemname\ extends KG-augmented LLM reasoning to this setting by constructing a 3D scene KG from multi-frame camera and LiDAR data, equipping it with ego-centric geometric and schema-constrained scene tools, and orchestrating them with an LLM agent to perform fine-grained, fact-grounded reasoning about autonomous-driving scenes.

%As shown in Table~\ref{tab:sota_comparison}, we summary the representative SOTA methods and their key properties. In a shot, classical 3D perception pipelines provide strong perception but lack explicit reasoning, whereas driving-oriented LLM frameworks offer reasoning but require heavy task-specific training and remain opaque. \systemname\ is designed to fill this gap by combining strong perception with explicit, explainable reasoning, without any task-specific finetuning of the LLM backbone.

%Taken together, existing 3D perception pipelines and driving-oriented LLM frameworks each cover only part of the requirements for reliable scene reasoning: they either focus on perception, rely on heavy task-specific training, or remain opaque in their reasoning process. As summarized in Table~\ref{tab:sota_comparison}, none of them simultaneously provide strong perception and reasoning ability, avoid task-specific finetuning, and support explainable reasoning. \systemname\ is designed to fill this gap.

\section{Conclusion}

In this paper, we study fine-grained factual reasoning in autonomous driving. 
We present \systemname, the first KG-based LLM reasoning framework for fine-grained question answering over 3D driving scenes. By coupling reliable scene fact construction with an LLM agent that performs explicit reasoning, \systemname\ produces answers grounded in verifiable scene facts rather than end-to-end black-box generation.
Experiments on two large-scale autonomous-driving QA benchmarks show that \systemname\ consistently outperforms prior perception and LLM-based methods, while substantially reducing hallucinations on challenging fine-grained reasoning tasks.
Overall, our results suggest that combining reliable scene fact construction with fact-grounded reasoning is a promising direction for more trustworthy autonomous-driving systems.

\section*{Acknowledgements}
This work has been funded in part by NSF, with award numbers \#2112665, \#2112167, \#2003279, \#2120019, \#2211386, \#2052809, \#1911095 and in part by PRISM and CoCoSys, centers in JUMP 2.0, an SRC program sponsored by DARPA.
%\newpage

%\begin{acks}
%This work was supported in part by National Science Foundation under Grants \#2003279, \#1826967, \#2100237, \#2112167, \#1911095, \#2112665, and in part by SRC under task \#3021.001. This work was also supported in part by PRISM and CoCoSys, centers in JUMP 2.0, an SRC program sponsored by DARPA.
%\end{acks}

\bibliographystyle{ACM-Reference-Format}
\bibliography{refs.bib}

@article{lecun2006tutorial,
  title={A tutorial on energy-based learning},
  author={LeCun, Yann and Chopra, Sumit and Hadsell, Raia and Ranzato, M and Huang, Fujie and others},
  journal={Predicting structured data},
  volume={1},
  number={0},
  year={2006}
}

@article{xu2024energy,
  title={Energy-based concept bottleneck models: Unifying prediction, concept intervention, and probabilistic interpretations},
  author={Xu, Xinyue and Qin, Yi and Mi, Lu and Wang, Hao and Li, Xiaomeng},
  journal={arXiv preprint arXiv:2401.14142},
  year={2024}
}

@inproceedings{liu2024ray,
  title={Ray denoising: Depth-aware hard negative sampling for multi-view 3d object detection},
  author={Liu, Feng and Huang, Tengteng and Zhang, Qianjing and Yao, Haotian and Zhang, Chi and Wan, Fang and Ye, Qixiang and Zhou, Yanzhao},
  booktitle={European Conference on Computer Vision},
  pages={200--217},
  year={2024},
  organization={Springer}
}

@inproceedings{chen2023focalformer3d,
  title={Focalformer3d: focusing on hard instance for 3d object detection},
  author={Chen, Yilun and Yu, Zhiding and Chen, Yukang and Lan, Shiyi and Anandkumar, Anima and Jia, Jiaya and Alvarez, Jose M},
  booktitle={Proceedings of the IEEE/CVF international conference on computer vision},
  pages={8394--8405},
  year={2023}
}

@article{peng2024multimath,
  title={Multimath: Bridging visual and mathematical reasoning for large language models},
  author={Peng, Shuai and Fu, Di and Gao, Liangcai and Zhong, Xiuqin and Fu, Hongguang and Tang, Zhi},
  journal={arXiv preprint arXiv:2409.00147},
  year={2024}
}

@article{wang2024measuring,
  title={Measuring multimodal mathematical reasoning with math-vision dataset},
  author={Wang, Ke and Pan, Junting and Shi, Weikang and Lu, Zimu and Ren, Houxing and Zhou, Aojun and Zhan, Mingjie and Li, Hongsheng},
  journal={Advances in Neural Information Processing Systems},
  volume={37},
  pages={95095--95169},
  year={2024}
}

@article{nishat2025aligning,
  title={Aligning Knowledge Graphs and Language Models for Factual Accuracy},
  author={Nishat, Nur A Zarin and Coletta, Andrea and Bellomarini, Luigi and Amouzouvi, Kossi and Lehmann, Jens and Vahdati, Sahar},
  journal={arXiv preprint arXiv:2507.13411},
  year={2025}
}

@inproceedings{ye2024correcting,
  title={Correcting Factual Errors in LLMs via Inference Paths Based on Knowledge Graph},
  author={Ye, Weiqi and Zhang, Qiang and Zhou, Xian and Hu, Wenpeng and Tian, Changhai and Cheng, Jiajun},
  booktitle={2024 International Conference on Computational Linguistics and Natural Language Processing (CLNLP)},
  pages={12--16},
  year={2024},
  organization={IEEE}
}

@article{ma2025knowledge,
  title={Knowledge graph-based retrieval-augmented generation for schema matching},
  author={Ma, Chuangtao and Chakrabarti, Sriom and Khan, Arijit and Moln{\'a}r, B{\'a}lint},
  journal={arXiv preprint arXiv:2501.08686},
  year={2025}
}

@inproceedings{agrawal2024can,
  title={Can knowledge graphs reduce hallucinations in llms?: A survey},
  author={Agrawal, Garima and Kumarage, Tharindu and Alghamdi, Zeyad and Liu, Huan},
  booktitle={Proceedings of the 2024 Conference of the North American Chapter of the Association for Computational Linguistics: Human Language Technologies (Volume 1: Long Papers)},
  pages={3947--3960},
  year={2024}
}

@article{xu2025vs,
  title={VS-Bench: Evaluating VLMs for Strategic Reasoning and Decision-Making in Multi-Agent Environments},
  author={Xu, Zelai and Xu, Zhexuan and Yi, Xiangmin and Yuan, Huining and Chen, Xinlei and Wu, Yi and Yu, Chao and Wang, Yu},
  journal={arXiv preprint arXiv:2506.02387},
  year={2025}
}

@inproceedings{li2025fine,
  title={Fine-grained evaluation of large vision-language models in autonomous driving},
  author={Li, Yue and Tian, Meng and Lin, Zhenyu and Zhu, Jiangtong and Zhu, Dechang and Liu, Haiqiang and Zhang, Yueyi and Xiong, Zhiwei and Zhao, Xinhai},
  booktitle={Proceedings of the IEEE/CVF International Conference on Computer Vision},
  pages={9431--9442},
  year={2025}
}

@article{wang2024mitigating,
  title={Mitigating hallucinations in large vision-language models with instruction contrastive decoding},
  author={Wang, Xintong and Pan, Jingheng and Ding, Liang and Biemann, Chris},
  journal={arXiv preprint arXiv:2403.18715},
  year={2024}
}

@article{liu2024survey,
  title={A survey on hallucination in large vision-language models},
  author={Liu, Hanchao and Xue, Wenyuan and Chen, Yifei and Chen, Dapeng and Zhao, Xiutian and Wang, Ke and Hou, Liping and Li, Rongjun and Peng, Wei},
  journal={arXiv preprint arXiv:2402.00253},
  year={2024}
}

@article{shi2025motion,
  title={Motion forecasting for autonomous vehicles: a survey},
  author={Shi, Jianxin and Chen, Jinhao and Wang, Yuandong and Sun, Li and Liu, Chunyang and Xiong, Wei and Wo, Tianyu},
  journal={arXiv preprint arXiv:2502.08664},
  year={2025}
}

@article{xia2024survey,
  title={A survey of autonomous vehicle behaviors: Trajectory planning algorithms, sensed collision risks, and user expectations},
  author={Xia, Taokai and Chen, Hui},
  journal={Sensors},
  volume={24},
  number={15},
  pages={4808},
  year={2024},
  publisher={MDPI}
}

@article{zhao2025survey,
  title={A survey of autonomous driving from a deep learning perspective},
  author={Zhao, Jingyuan and Wu, Yuyan and Deng, Rui and Xu, Susu and Gao, Jinpeng and Burke, Andrew},
  journal={ACM Computing Surveys},
  volume={57},
  number={10},
  pages={1--60},
  year={2025},
  publisher={ACM New York, NY}
}

@inproceedings{Liu_2024_CVPR,
  author    = {Haotian Liu and Chunyuan Li and Yuheng Li and Yong Jae Lee},
  title     = {Improved Baselines with Visual Instruction Tuning},
  booktitle = {Proceedings of the IEEE/CVF Conference on Computer Vision and Pattern Recognition (CVPR)},
  year      = {2024},
  pages     = {26296--26306}
}

@misc{qwen3technicalreport,
  title         = {Qwen3 Technical Report},
  author        = {Qwen Team},
  year          = {2025},
  eprint        = {2505.09388},
  archivePrefix = {arXiv},
  primaryClass  = {cs.CL},
  url           = {https://arxiv.org/abs/2505.09388}
}

@inproceedings{yu2019deep,
  title={Deep modular co-attention networks for visual question answering},
  author={Yu, Zhou and Yu, Jun and Cui, Yuhao and Tao, Dacheng and Tian, Qi},
  booktitle={Proceedings of the IEEE/CVF conference on computer vision and pattern recognition},
  pages={6281--6290},
  year={2019}
}

@article{wang2025internvl3_5,
  title   = {InternVL3.5: Advancing Open-Source Multimodal Models in Versatility, Reasoning, and Efficiency},
  author  = {Wang, Weiyun and Gao, Zhangwei and Gu, Lixin and Pu, Hengjun and Cui, Long and Wei, Xingguang and Liu, Zhaoyang and Jing, Linglin and Ye, Shenglong and Shao, Jie and others},
  journal = {arXiv preprint arXiv:2508.18265},
  year    = {2025}
}

@inproceedings{yin2024fusion,
  title={Is-fusion: Instance-scene collaborative fusion for multimodal 3d object detection},
  author={Yin, Junbo and Shen, Jianbing and Chen, Runnan and Li, Wei and Yang, Ruigang and Frossard, Pascal and Wang, Wenguan},
  booktitle={Proceedings of the IEEE/CVF conference on computer vision and pattern recognition},
  pages={14905--14915},
  year={2024}
}

@online{graphrag-msr,
  author  = {Darren Edge and Ha Trinh and Steven Truitt and Jonathan Larson},
  title   = {GraphRAG: New tool for complex data discovery now on GitHub},
  date    = {2024-07-02},
  url     = {https://www.microsoft.com/en-us/research/blog/graphrag-new-tool-for-complex-data-discovery-now-on-github/},
  urldate = {2025-11-04},
  organization = {Microsoft Research},
}

@article{liu2024knowledge,
  title={Knowledge graph-enhanced large language models via path selection},
  author={Liu, Haochen and Wang, Song and Zhu, Yaochen and Dong, Yushun and Li, Jundong},
  journal={arXiv preprint arXiv:2406.13862},
  year={2024}
}

@article{zhang2024knowledge,
  title={Knowledge graph enhanced large language model editing},
  author={Zhang, Mengqi and Ye, Xiaotian and Liu, Qiang and Ren, Pengjie and Wu, Shu and Chen, Zhumin},
  journal={arXiv preprint arXiv:2402.13593},
  year={2024}
}

@article{coppolillo2025injecting,
  title={Injecting Knowledge Graphs into Large Language Models},
  author={Coppolillo, Erica},
  journal={arXiv preprint arXiv:2505.07554},
  year={2025}
}

@article{sun2023think,
  title={Think-on-graph: Deep and responsible reasoning of large language model on knowledge graph},
  author={Sun, Jiashuo and Xu, Chengjin and Tang, Lumingyuan and Wang, Saizhuo and Lin, Chen and Gong, Yeyun and Ni, Lionel M and Shum, Heung-Yeung and Guo, Jian},
  journal={arXiv preprint arXiv:2307.07697},
  year={2023}
}

@inproceedings{zhao2024kg,
  title={Kg-cot: Chain-of-thought prompting of large language models over knowledge graphs for knowledge-aware question answering},
  author={Zhao, Ruilin and Zhao, Feng and Wang, Long and Wang, Xianzhi and Xu, Guandong},
  booktitle={Proceedings of the Thirty-Third International Joint Conference on Artificial Intelligence (IJCAI-24)},
  pages={6642--6650},
  year={2024},
  organization={International Joint Conferences on Artificial Intelligence}
}

@article{li2024simple,
  title={Simple is effective: The roles of graphs and large language models in knowledge-graph-based retrieval-augmented generation},
  author={Li, Mufei and Miao, Siqi and Li, Pan},
  journal={arXiv preprint arXiv:2410.20724},
  year={2024}
}

@article{huang2021bevdet,
  title={Bevdet: High-performance multi-camera 3d object detection in bird-eye-view},
  author={Huang, Junjie and Huang, Guan and Zhu, Zheng and Ye, Yun and Du, Dalong},
  journal={arXiv preprint arXiv:2112.11790},
  year={2021}
}

@article{li2022bevformer,
  title={Bevformer: Learning bird’s-eye-view representation from multi-camera images via spatiotemporal transformers.(2022)},
  author={Li, Zhiqi and Wang, Wenhai and Li, Hongyang and Xie, Enze and Sima, Chonghao and Lu, Tong and Yu, Qiao and Dai, Jifeng},
  journal={URL https://arxiv. org/abs/2203.17270},
  year={2022}
}

@inproceedings{li2024bevnext,
  title={Bevnext: Reviving dense bev frameworks for 3d object detection},
  author={Li, Zhenxin and Lan, Shiyi and Alvarez, Jose M and Wu, Zuxuan},
  booktitle={Proceedings of the IEEE/CVF conference on computer vision and pattern recognition},
  pages={20113--20123},
  year={2024}
}

@inproceedings{yin2021center,
  title={Center-based 3d object detection and tracking},
  author={Yin, Tianwei and Zhou, Xingyi and Krahenbuhl, Philipp},
  booktitle={Proceedings of the IEEE/CVF conference on computer vision and pattern recognition},
  pages={11784--11793},
  year={2021}
}

@inproceedings{zhou2022centerformer,
  title={Centerformer: Center-based transformer for 3d object detection},
  author={Zhou, Zixiang and Zhao, Xiangchen and Wang, Yu and Wang, Panqu and Foroosh, Hassan},
  booktitle={European Conference on Computer Vision},
  pages={496--513},
  year={2022},
  organization={Springer}
}

@inproceedings{chae2024lidar,
  title={Lidar-based all-weather 3d object detection via prompting and distilling 4d radar},
  author={Chae, Yujeong and Kim, Hyeonseong and Oh, Changgyoon and Kim, Minseok and Yoon, Kuk-Jin},
  booktitle={European Conference on Computer Vision},
  pages={368--385},
  year={2024},
  organization={Springer}
}

@article{liang2022bevfusion,
  title={Bevfusion: A simple and robust lidar-camera fusion framework},
  author={Liang, Tingting and Xie, Hongwei and Yu, Kaicheng and Xia, Zhongyu and Lin, Zhiwei and Wang, Yongtao and Tang, Tao and Wang, Bing and Tang, Zhi},
  journal={Advances in Neural Information Processing Systems},
  volume={35},
  pages={10421--10434},
  year={2022}
}

@inproceedings{jiao2023msmdfusion,
  title={Msmdfusion: Fusing lidar and camera at multiple scales with multi-depth seeds for 3d object detection},
  author={Jiao, Yang and Jie, Zequn and Chen, Shaoxiang and Chen, Jingjing and Ma, Lin and Jiang, Yu-Gang},
  booktitle={Proceedings of the IEEE/CVF conference on computer vision and pattern recognition},
  pages={21643--21652},
  year={2023}
}

@article{xu2024drivegpt4,
  title={Drivegpt4: Interpretable end-to-end autonomous driving via large language model},
  author={Xu, Zhenhua and Zhang, Yujia and Xie, Enze and Zhao, Zhen and Guo, Yong and Wong, Kwan-Yee K and Li, Zhenguo and Zhao, Hengshuang},
  journal={IEEE Robotics and Automation Letters},
  year={2024},
  publisher={IEEE}
}

@article{mao2023gpt,
  title={Gpt-driver: Learning to drive with gpt},
  author={Mao, Jiageng and Qian, Yuxi and Ye, Junjie and Zhao, Hang and Wang, Yue},
  journal={arXiv preprint arXiv:2310.01415},
  year={2023}
}

@inproceedings{shao2024lmdrive,
  title={Lmdrive: Closed-loop end-to-end driving with large language models},
  author={Shao, Hao and Hu, Yuxuan and Wang, Letian and Song, Guanglu and Waslander, Steven L and Liu, Yu and Li, Hongsheng},
  booktitle={Proceedings of the IEEE/CVF Conference on Computer Vision and Pattern Recognition},
  pages={15120--15130},
  year={2024}
}

@article{tian2024drivevlm,
  title={Drivevlm: The convergence of autonomous driving and large vision-language models},
  author={Tian, Xiaoyu and Gu, Junru and Li, Bailin and Liu, Yicheng and Wang, Yang and Zhao, Zhiyong and Zhan, Kun and Jia, Peng and Lang, Xianpeng and Zhao, Hang},
  journal={arXiv preprint arXiv:2402.12289},
  year={2024}
}

@article{mao2023language,
  title={A language agent for autonomous driving},
  author={Mao, Jiageng and Ye, Junjie and Qian, Yuxi and Pavone, Marco and Wang, Yue},
  journal={arXiv preprint arXiv:2311.10813},
  year={2023}
}

@inproceedings{yang2025lidar,
  title={Lidar-llm: Exploring the potential of large language models for 3d lidar understanding},
  author={Yang, Senqiao and Liu, Jiaming and Zhang, Renrui and Pan, Mingjie and Guo, Ziyu and Li, Xiaoqi and Chen, Zehui and Gao, Peng and Li, Hongsheng and Guo, Yandong and others},
  booktitle={Proceedings of the AAAI Conference on Artificial Intelligence},
  volume={39},
  number={9},
  pages={9247--9255},
  year={2025}
}

@inproceedings{marcu2024lingoqa,
  title={Lingoqa: Visual question answering for autonomous driving},
  author={Marcu, Ana-Maria and Chen, Long and H{\"u}nermann, Jan and Karnsund, Alice and Hanotte, Benoit and Chidananda, Prajwal and Nair, Saurabh and Badrinarayanan, Vijay and Kendall, Alex and Shotton, Jamie and others},
  booktitle={European Conference on Computer Vision},
  pages={252--269},
  year={2024},
  organization={Springer}
}

@inproceedings{yu2025crema,
  title={CREMA: Generalizable and Efficient Video-language Reasoning via Multimodal Modular Fusion},
  author={Yu, S and Yoon, J and Bansal, M},
  year={2025},
  organization={Proceedings of the International Conference on Learning Representations}
}

@inproceedings{sima2024drivelm,
  title={Drivelm: Driving with graph visual question answering},
  author={Sima, Chonghao and Renz, Katrin and Chitta, Kashyap and Chen, Li and Zhang, Hanxue and Xie, Chengen and Bei{\ss}wenger, Jens and Luo, Ping and Geiger, Andreas and Li, Hongyang},
  booktitle={European conference on computer vision},
  pages={256--274},
  year={2024},
  organization={Springer}
}

@inproceedings{cao2024maplm,
  title={Maplm: A real-world large-scale vision-language benchmark for map and traffic scene understanding},
  author={Cao, Xu and Zhou, Tong and Ma, Yunsheng and Ye, Wenqian and Cui, Can and Tang, Kun and Cao, Zhipeng and Liang, Kaizhao and Wang, Ziran and Rehg, James M and others},
  booktitle={Proceedings of the IEEE/CVF conference on computer vision and pattern recognition},
  pages={21819--21830},
  year={2024}
}

@inproceedings{qian2024nuscenes,
  title={Nuscenes-qa: A multi-modal visual question answering benchmark for autonomous driving scenario},
  author={Qian, Tianwen and Chen, Jingjing and Zhuo, Linhai and Jiao, Yang and Jiang, Yu-Gang},
  booktitle={Proceedings of the AAAI Conference on Artificial Intelligence},
  volume={38},
  number={5},
  pages={4542--4550},
  year={2024}
}

\end{document}